\documentclass[10pt,twocolumn,letterpaper]{article}

\usepackage{iccv}
\usepackage{times}
\usepackage{epsfig}
\usepackage{graphicx}
\usepackage{amsmath}
\usepackage{amssymb}

\usepackage{esvect}
\usepackage{booktabs}
\usepackage{multirow}
\usepackage{comment}
\usepackage{subcaption}
\usepackage{footnote}
\makesavenoteenv{table*}

\usepackage[dvipsnames]{xcolor}

\usepackage[pagebackref=true,breaklinks=true,letterpaper=true,colorlinks,bookmarks=false]{hyperref}

\iccvfinalcopy 

\def\iccvPaperID{5172} 

\newcommand{\LT}[1]{\textcolor{blue}{[LT: #1]}} 

\ificcvfinal\fi
\begin{document}

\title{SCSampler: Sampling Salient Clips from Video for Efficient Action Recognition}

\author{Bruno Korbar \qquad
Du Tran \qquad
Lorenzo Torresani\vspace{8pt}\\
Facebook AI\\
{\tt\small \{bkorbar, trandu, torresani\}@fb.com}
}

\maketitle

\begin{abstract}
While many action recognition datasets consist of collections of brief, trimmed videos each containing a relevant action, videos in the real-world (e.g., on YouTube) exhibit very different properties: they are often several minutes long, where brief relevant clips are often interleaved with segments of extended duration containing little change. Applying densely an action recognition system to every temporal clip within such videos is prohibitively expensive. Furthermore, as we show in our experiments, this results in suboptimal recognition accuracy as informative predictions from relevant clips are outnumbered by meaningless classification outputs over long uninformative sections of the video. In this paper we introduce a lightweight ``clip-sampling'' model that can efficiently identify the most salient temporal clips within a long video. We demonstrate that the computational cost of action recognition on untrimmed videos can be dramatically reduced by invoking recognition only on these most salient clips. Furthermore, we show that this yields significant gains in recognition accuracy compared to analysis of all clips or randomly/uniformly selected clips. On Sports1M, our clip sampling scheme elevates the accuracy of an already state-of-the-art action classifier by $7\%$ and reduces by more than 15 times its computational cost.
\end{abstract}

\section{Introduction}

Most modern action recognition models operate by applying a deep CNN over clips of fixed temporal length~\cite{TwoStreamAZ:NIPS14,carreira2017quo,tran2017closer,NonLocal,SlowFast}. Video-level classification is obtained by aggregating the clip-level predictions over the entire video, either in the form of simple averaging or by means of more sophisticated schemes modeling temporal structure~\cite{NgEtAl:CVPR15, VarolEtAl:TPAMI17, Ghirdar:CVPR17}. Scoring a clip classifier densely over the entire sequence is a reasonable approach for short videos. However, it becomes computationally impractical for real-world videos that may be up to an hour long, such as some of the sequences in the Sports1M dataset~\cite{Sports1M}. In addition to the issue of computational cost, long videos often include segments of extended duration that provide irrelevant information for the recognition of the action class. Pooling information from all clips without consideration of their relevance may cause poor video-level classification, as informative clip predictions are outnumbered by uninformative predictions over long unimportant segments.

\begin{figure}[t]
\captionsetup{font=small}
\small
\centering
\begin{subfigure}[t]{0.45\columnwidth}
  \centering
  \includegraphics[width=.8\linewidth]{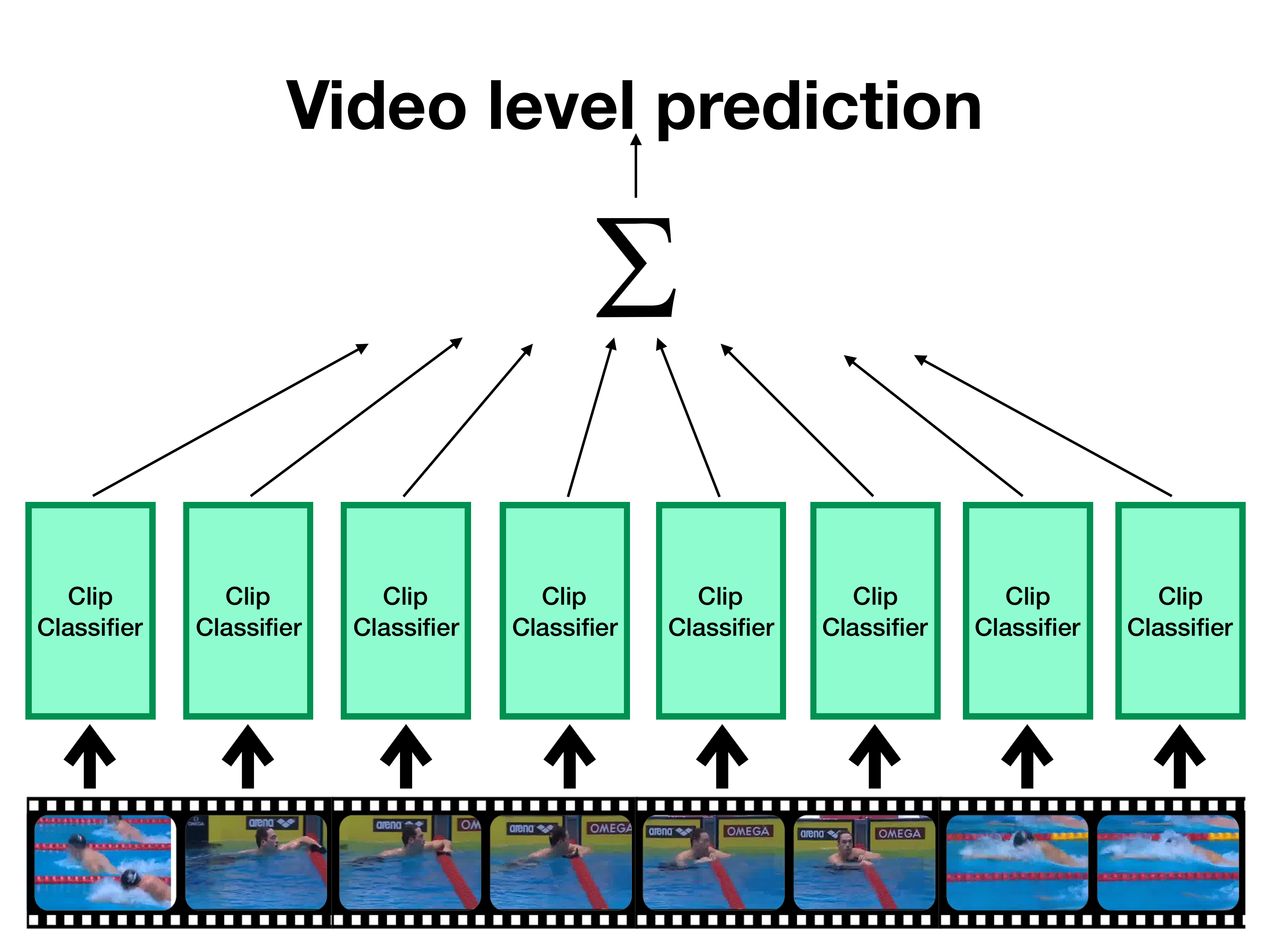}
  \caption{Dense predictions}
  \label{fig:sub1}
\end{subfigure}%
\begin{subfigure}[t]{.48\columnwidth}
  \centering
  \includegraphics[width=.8\linewidth]{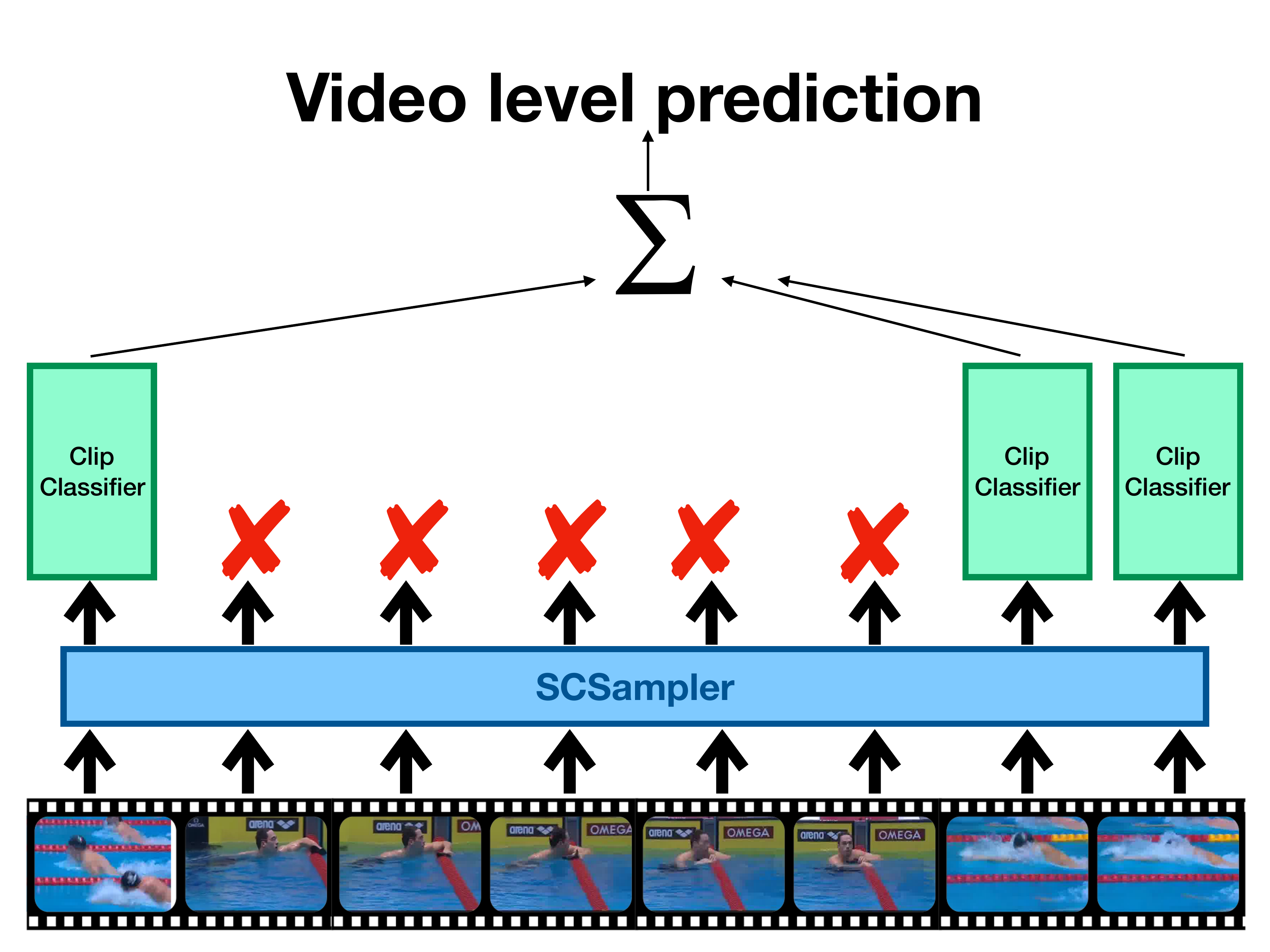}
  \caption{Our suggested approach}
  \label{fig:sub2}
\end{subfigure}
\caption{\small Overview: video-level classification by averaging (a) dense clip-level predictions vs (b) selected predictions computed only for salient clips. SCSampler yields accuracy gains and runtime speedups by eliminating predictions over uninformative clips.}
\label{fig:overview}
\vspace{-8pt}

\end{figure}

In this work we propose a simple scheme to address these problems (see Fig.~\ref{fig:overview} for a high-level illustration of the approach). It consists in training an extremely lightweight network to determine the saliency of a candidate clip. Because the computational cost of this network is more than one order of magnitude lower than the cost of existing 3D CNNs for action recognition~\cite{carreira2017quo,tran2017closer}, it can be evaluated efficiently over all clips of even long videos. We refer to our network as {\em SCSampler} (Salient Clip Sampler), as it samples a reduced set of salient clips from the video for analysis by the action classifier. We demonstrate that restricting the costly action classifier to run only on the clips identified as the most salient by SCSampler, yields not only significant savings in runtime but also large improvements in video classification accuracy: on Sports1M our scheme yields a speedup of 15$\times$ and an accuracy gain of $7\%$ over an already state-of-the-art classifier. 

Efficiency is a critical requirement in the design of SCSampler. We present two main variants of our sampler. The first operates directly on compressed video~\cite{R1:MR1,wu2018coviar,zhang2016cvpr}, thus eliminating the need for costly decoding. The second looks only at the audio channel, which is low-dimensional and can therefore be processed very efficiently. As in recent multimedia work~\cite{arandjelovic2017look,aytar2016soundnet,Gao18ECCV,owens2018audio}, our audio-based sampler exploits the inherent semantic correlation between the audio and the visual elements of a video. 
We also show that combining our video-based sampler with the audio-based sampler leads to further gains in recognition accuracy.

We propose and evaluate two distinct learning objectives for salient clip sampling. One of them trains the sampler to operate optimally with the given clip classifier, while the second formulation is classifier-independent. We show that, in some settings, the former leads to improved accuracy, while the benefit of the latter is that it can be used without retraining with any clip classifier, making this model a general and powerful off-the-shelf tool to improve both the runtime and the accuracy of clip-based action classification. Finally, we show that although our sampler is trained over specific action classes in the training set, its benefits extend even to recognition of novel action classes.


\section{Related work}

The problem of selecting relevant frames, clips or segments within a video has been investigated for various applications. For example, video summarization~\cite{GongEtAl:NIPS16, GygliEtAl:CVPR15,  MahasseniEtAl:CVPR17, PotapovEtAl:ECCV14, zhang2016cvpr,  ZhangEtAl:CVPR16, ZhangEtAl:ECCV16} and the automatic production of sport highlights~\cite{MerlerEtAl:CVPRW18, MerlerEtAl:TMM18} entail creating a much shorter version of the original video by concatenating a small set of snippets corresponding to the most informative or exciting moments. 
The aim of these systems is to generate a video composite that is pleasing and compelling for the user. 
Instead the objective of our model is to select a set of segments of fixed duration (i.e., clips) so as to make video-level classification as accurate and as unambiguous as possible. 

More closely related to our task is the problem of action localization~\cite{Mihir:CVPR14, ShouEtAL:CVPR16, ShouEtAl:CVPR17, xu2017rc3d, ZhaoEtAl:ICCV17}, where the objective is to localize the temporal start and end of each action within a given untrimmed video and to recognize the action class. Action localization is often approached through a two-step mechanism~\cite{ BuchEtAl:CVPR17, EscorciaEtal:ECCV16, BuchEtAl:CVPR17, GaoEtAl:ICCV17, Gao18ECCV, HeilbronEtAl:CVPR16, LinEtAl:ECCV18, ActionSearch}, where first an action proposal method identifies candidate action segments, and then a more sophisticated approach validates the class of each candidate and refines its temporal boundaries. Our framework is reminiscent of this two-step solution, as our sampler can be viewed as selecting candidate clips for accurate evaluation by the action classifier. However, several key differences exist between our objective and that of action localization. Our system is aimed at video classification, where the assumption is that each video contains a single action class. Action proposal methods solve the harder problem of finding segments of different lengths and potentially belonging to different classes within the input video.
While in action localization the validation model is typically trained using the candidate segments produced by the proposal method, the opposite is true in our scenario: the sampler is learned for a given pretrained clip classifier, which is left unmodified by our approach. Finally, the most fundamental difference is that high efficiency is a critical requirement in the design of our clip sampler. Our sampler must be orders of magnitude faster than the clip classifier to make our approach worthwhile. Conversely, most action proposal or localization methods are based on optical flow~\cite{LinEtAl:CVPRW17, LinEtAl:ECCV18} or deep action-classifier features~\cite{BuchEtAl:CVPR17, Gao18ECCV, xu2017rc3d} that are typically at least as expensive to compute as the output of a clip classifier. For example, the TURN TAP system~\cite{GaoEtAl:ICCV17} is one of the fastest existing action proposal methods and yet, its computational cost exceeds by more than one order of magnitude that of our scheme. For 60 seconds of untrimmed video, TURN TAP has a cost of 4128 GFLOPS; running densely our clip classifier (MC3-18~\cite{tran2017closer}) over the 60 seconds would actually cost less, at 1097 GFLOPs; our sampling scheme lowers the cost down dramatically, to only 168 GFLOPs. 

Closer to our intent are methods that remove from consideration uninformative sections of the video. This is typically achieved by means of temporal models that ``skip'' segments by leveraging past observations to predict which future frames to consider next~\cite{yeung2016end, fan18ijcai, Wu_2019_CVPR}. Instead of learning to skip, our approach relies on a fast sampling procedures that evaluates all segments in a video and then performs further analysis on the most salient ones.

Our approach belongs to the genre of work that performs video classification by aggregating temporal information from long videos~\cite{GaidonEtAl:TPAMI2013, R1:MR3, NgEtAl:CVPR15, PirsiavashEtAl:CVPR14,VarolEtAl:TPAMI17, WangCherian:ECCV18,  WangEtaAl:IJCV16, WangEtAl_TSN:ECCV16, WangEtAl:CVPR16, WuEtAl:CVPR19, ZhouEtAl:ECCV18}. Our aggregation scheme is very simple, as it merely averages the scores of action classifiers over the selected clips. Yet, we note that the most recent state-of-the-art action classifiers operate precisely under this simple scheme. Examples include Two-Stream Networks~\cite{TwoStreamAZ:NIPS14}, I3D~\cite{carreira2017quo}, R(2+1)D~\cite{tran2017closer}, Non-Local Networks~\cite{NonLocal}, SlowFast~\cite{SlowFast}. While in these prior studies clips are sampled densely or at random, our experiment suggest that our sampling strategy yields significant gains in accuracy over both dense, random, and uniform sampling and it is as fast as random sampling.


\section{Technical approach}
\label{techapproach}
Our approach consists in extracting a small set of relevant clips from a video by scoring densely each clip with a lightweight saliency model. We refer to this model as the ``sampler'' since it is used to sample clips from the video. We formally define the task in subsection~\ref{subs:prob}, proceed to present two different learning objectives for the sampler in section~\ref{subs:loss_choice}, and finally discuss sampler architecture choices and features in subsection~\ref{sec:sampler}.

\subsection{Problem Formulation}
\label{subs:prob}

\noindent{\bf Video classification from clip-level predictions.}
We assume we are given a pretrained action classifier $\text{{\bf f}}: \mathbb{R}^{F \times 3 \times H \times W} \rightarrow [0, 1]^C$ operating on short, fixed-length clips of $F$ RGB frames with spatial resolution $H \times W$ and producing output classification probabilities over a set of action classes $\{1, \hdots, C\}$. We note that most modern action recognition systems~\cite{carreira2017quo, feichtenhofer2016spatiotemporal, tran2015learning,  tran2017closer} fall under this model and, typically, they constrain the number of frames $F$ to span just a handful of seconds in order to keep memory consumption manageable during training and testing. Given a test video $v \in \mathbb{R}^{T \times 3 \times H \times W}$ of arbitrary length $T$, video-level classification through the clip-classifier $\text{{\bf f}}$ is achieved by first splitting the video $v$ into a set of clips $\{v^{(i)}\}_{i=1}^{L}$ with each clip $v^{(i)} \in \mathbb{R}^{F \times 3 \times H \times W}$ consisting of $F$ adjacent frames and where $L$ denotes the total number of clips in the video. The splitting is usually done by taking clips every $F$ frames in order to have a set of non-overlapping clips that spans the entirety of the video. A final video-level prediction is then computed by aggregating the individual clip-level predictions. In other words, if we denote with $\text{\em{aggr}}$ the aggregation operator, the video-level classifier $\hat{\bf f}$ is obtained as $\hat{{\bf f}}(v) = \text{\em{aggr}}(\{ {\bf f}(v^{(i)}) \}_{i=1}^{L})$.

\noindent Most often, the aggregator is a simple pooling operator which averages the individual clip scores (i.e., $\hat{\bf f}(v)  = 1/L \sum_{i=1}^L {\bf f}(v^{(i)})$)~\cite{carreira2017quo, SlowFast,TwoStreamAZ:NIPS14,tran2017closer,NonLocal} but more sophisticated schemes based on RNNs~\cite{yue2015beyond} have also been employed. 

\noindent\textbf{Video classification from selected clips} In this paper we are interested in scenarios where the videos $v$ are untrimmed and may be  quite long. In such cases, applying the clip classifier ${\bf f}$ to every clip will result in a very large inference cost. Furthermore, aggregating predictions from the entire video may produce poor action recognition accuracy since in long videos the target action is unlikely to be exhibited in every clip. Thus, our objective is to design a method that can {\em efficiently} identify a subset $\mathcal{S}(v; K)$ of $K$ salient clips in the video (i.e., $\mathcal{S}(v; K) \in 2^{\{1,\hdots,L\}}$ with $|\mathcal{S}(v; K)|=K$) and to reduce video-level prediction to be computed from this set of $K$ clip-level predictions as $\hat{f}_{\mathcal{S}(v; K)}(v) = \text{\em{aggr}}(\{{\bf f} (v^{(i)})\}_{i \in \mathcal{S}(v; K)})$ ($K$ is hyper-parameter studied in our experiments). By constraining the application of the costly classifier ${\bf f}$ to only $K$ clips, inference will be efficient even on long videos. Furthermore, by making sure that $\mathcal{S}(v; K)$ includes a sample of the most salient clips in $v$, recognition accuracy may improve as irrelevant or ambiguous clips will be discarded from consideration and will be prevented from polluting the video-level prediction. We note that in this work we address the problem of clip selection for a given pretrained clip classifier ${\bf f}$, which is left unmodified by our method. This renders our approach useful as a post-training procedure to further improve performance of existing classifiers both in terms of inference speed as well as recognition accuracy.

\noindent{\bf Our clip sampler.}  In order to achieve our goal we propose a simple solution that consists in learning a highly efficient clip-level saliency model $s(.)$ that provides for each clip in the video a ``saliency score'' in $[0, 1]$. Specifically, our saliency model $s(.)$ takes as input clip features $\phi^{(i)} = \phi(v^{(i)}) \in \mathbb{R}^d$ that are fast to compute from the raw clip $v^{(i)}$ and that have low dimensionality ($d$) so that each clip can be analyzed very efficiently. The saliency model $s: \mathbb{R}^{d} \rightarrow [0, 1]$  is designed to be orders of magnitude faster than ${\bf f}$, thus enabling the possibility to score $s$ on every single clip of the video to find the $K$ most salient clips without adding any significant overhead. The set $\mathcal{S}(v; K)$ is then obtained as $\mathcal{S}(v; K) = \text{{\em top}}K( \{{s} (\phi^{(i)})\}_{i=1}^{L} )$ where $\text{{\em top}}K$ returns the indices of the top-$K$ values in the set. We show that evaluating {\bf f} on these selected set, i.e., computing $\hat{f}_{\mathcal{S}(v; K)}(v) = \text{\em{aggr}}(\{{\bf f} (v^{(i)})\}_{i \in \mathcal{S}(v; K)})$) results in significantly higher accuracy compared to aggregating clip-level prediction over all clips.

\noindent In order to learn the sampler $s$, we use a training set $\mathcal{D}$ of untrimmed video examples, each annotated with a label indicating the action performed in the video: $\mathcal{D} = \{(v_1, y_1), \hdots, (v_N, y_N)\}$ with $v_n \in \mathbb{R}^{T_n \times 3 \times H \times W }$ denoting the $n$-th video and $y_n \in \{1, \hdots, C\}$ indicating its action label. In our experiments, we use as training set $\mathcal{D}$ the same set of examples that was used to train the clip classifier ${\bf f}$. This setup allows us to demonstrate that the gains in recognition accuracy are not due to leveraging additional data but instead are the result of learning to detect the most salient clips for ${\bf f}$ within each video. 

\noindent{\bf Oracle sampler.}  In this work we compare our sampler against an ``oracle'' $\mathcal{O}$ that makes use of the action label $y$ to select the best $K$ clips in the video for classification with ${\bf f}$. The oracle set is formally defined as $\mathcal{O}(v,y; K) = \text{{\em top}}K( \{{f}_y (v^{(i)})\}_{i=1}^{L} )$. Note that $\mathcal{O}$ is obtained by looking for the clips that yield the $K$ highest action classification scores for the {\em ground-truth} label $y$ under the costly action classifier ${\bf f}$. In real scenarios the oracle cannot be constructed as it requires knowing the true label and it involves dense application of ${\bf f}$ over the entire video, which defeats the purpose of the sampler. Nevertheless, in this work we use the oracle to obtain an upper bound on the accuracy of the sampler. Furthermore, we apply the oracle to the training set $\mathcal{D}$ to form pseudo ground-truth data to train our sampler, as discussed in the next subsection.


\subsection{Learning Objectives for SCSampler}
\label{subs:loss_choice}

We consider two choices of learning objective for the sampler and experimentally compare them in~\ref{sec:abl_loss}.

\subsubsection{Training the sampler as an action classifier}
\label{sec:acloss}

A na\"ive way to approach the learning of the sampler $s$ is to first train a lightweight action classifier ${\bf h}(\phi_n^{(i)}) \in [0, 1]^C$ on the training set $\mathcal{D}$ by forming clip examples $(\phi_n^{(i)}, y_n)$ using the low-dimensional clip features $\phi_n^{(i)} = \phi(v_n^{(i)}) \in \mathbb{R}^d$. Note that this is equivalent to assuming that every clip in the training video contains a manifestation of the target action. 
Then, given a new untrimmed test video $v$, we can compute the saliency score of a clip in the video as the maximum classification score over the $C$ classes, i.e., $s(\phi^{(i)}) = \max_{c \in \{1,\hdots,C\}} h_c(\phi^{(i)})$. The rationale behind this choice is that a salient clip is expected to elicit a strong response by the classifier, while irrelevant or ambiguous clips are likely to cause weak predictions for all classes. We refer to this variant of our loss as {\em AC} (Action Classification).

\subsubsection{Training the sampler as a saliency ranker}
\label{sec:salrankloss}

One drawback of {\em AC} is that the sampler is trained as an action classifier {\em independently} from the model ${\bf f}$ and by assuming that all clips are equally relevant. Instead, ideally we would like the sampler to select clips that are most useful to our given {\bf f}. To achieve this goal we propose to train the sampler to recognize the relative importance of the clips within a video with respect to the classification output of {\bf f} for the correct action label. To achieve this goal, we define pseudo ground-truth binary labels $z_n^{(i,j)}$ for pairs of clips $(i,j)$ from the same video $v_n$:
\begin{eqnarray}
z_n^{(i,j)} = \begin{cases} 
\begin{array}{cl}
1 & \text{ if~~ } f_{y_n}(v_n^{(i)}) > f_{y_n}(v_n^{(j)})\\
-1 & \text{ otherwise}
\end{array}
\end{cases}
\end{eqnarray}
We train $s$ by minimizing a ranking loss over these pairs:
\begin{equation} \label{eq:MRLoss}
   \ell(\phi_n^{(i)}, \phi_n^{(j)}) = \max\left( -z_n^{(i,j)} [s(\phi_n^{(i)})-s(\phi_n^{(j)})+\eta], 0\right)
\end{equation}
where $\eta$ is a margin hyper-parameter. This loss encourages the sampler to rank higher clips that produce a higher classification score under ${\bf f}$ for the correct label. We refer to this sampler loss as {\em SAL-RANK} (Saliency Ranking).

\subsection{Sampler Architecture}\label{sec:sampler}

Due to the tight runtime requirements, we restrict our sampler to operate on two types of features that can be computed efficiently from video and that yield a very compact representation to process. The first type of features are obtained directly from the compressed video {\em without} the need for decoding. Prior work has shown that features computed from compressed video can even be used for action recognition~\cite{wu2018coviar}. We describe in detail these features in subsection~\ref{subsubvisualsampler}. The second type of features are audio features, which are even more compact and faster to compute than the compressed video features. Recent work~\cite{ arandjelovic2017look, arandjelovic2017objects, aytar2016soundnet, Gao18ECCV, korbar2018cooperative, owens2018audio,  zhao2018sound} has shown that the audio channel provides strong cues about the content of the video and this semantic correlation can be leveraged for various applications.

In subsection~\ref{subsubaudiosampler} we discuss how we can exploit the low-dimensional audio modality to find efficiently salient clips in a video.

\subsubsection{Visual sampler}
\label{subsubvisualsampler}

Wu et al.~\cite{wu2018coviar}  recently introduced an accurate action recognition model directly trained on compressed video. Modern codecs such as MPEG-4 and H.264 represent video in highly compressed form by storing the information in a set of sparse {\em I-frames}, each followed by a sequence of {\em P-frames}. An I-frame (IF) represents the RGB-frame in a video just as an image. Each I-frame is followed by 11 P-frames, which encode the 11 subsequent frames in terms of motion displacement ({MD}), and RGB-residual ({RGB-R}). MDs capture the frame-to-frame 2D motion while RGB-Rs store the remaining difference in RGB values between adjacent frames {\em after} having applied the MD field to rewarp the frame. In~\cite{wu2018coviar} it was shown that each of these three modalities (IFs, MDs, RGB-Rs) provides useful information for efficient and accurate action recognition in video. Inspired by this prior work, here we train three separate ResNet-18 networks~\cite{he2016residual} on these three inputs as samplers using the learning objectives outlined in the previous subsection. The first ResNet-18 takes as input an IF of size $H\times W\times 3$. The second is trained on MD frames, which have size $H/16 \times W/16 \times 2$: the 2 channels encode the horizontal and vertical motion displacements at a resolution that is 16 times smaller than the original video. The third ResNet-18 is fed individual RGB-Rs of size $H \times W \times 3$. At test time we average the predictions of these 3 models over all the I-frames and P-frames (MDs and RGB-Rs) within the clip to obtain a final global saliency score for the clip. As an alternative to ResNet-18, we experimented also with a lightweight ShuffleNet architecture~\cite{zhang2018shufflenet} of 26 layers. We compare these models in~\ref{sec:abl_archfeat}. We do not present results for the large ResNet-152 model that was used in~\cite{wu2018coviar}, since it adds a cost of 3 GFLOPS per clip which far exceeds the computational budget of our application.

\subsubsection{Audio sampler}
\label{subsubaudiosampler}

We model our audio sampler after the VGG-like audio networks used in~\cite{chung2016out, arandjelovic2017look, korbar2018cooperative}. Specifically, we first extract MEL-spectrograms from audio segments twice as as long as the video-clips, but with stride equal to the video-clip length. This stride is chosen to obtain an audio-based saliency score for every video clip used by the action recognizer {\bf f}. However, for the audio sampler we use an observation window twice as long as the video clip since we found this to yield better results. A series of 200 time samples is taken within each audio segment and processed using $40$ MEL filters. This yields a descriptor of size $40 \times 200$. This representation is compact and can be analyzed efficiently by the sampler. We treat this descriptor as an image and process it using a VGG network~\cite{vgg2014} of 18 layers. The details of the architecture are given in the supplementary material.

\subsubsection{Combining video and audio saliency}\label{subsubcombined}

Since audio and video provide correlated but distinct cues, we investigated several schemes for combining the saliency predictions from these two modalities. With {\em AV-convex-score} we denote a model that simply combines the audio-based score $s^A(v^{(i)})$ and the video-based score $s^V(v^{(i)})$ by means of a convex combination $\alpha s^V(v^{(i)}) + (1-\alpha) s^A(v^{(i)})$ where $\alpha$ is a scalar hyperparameter. The scheme {\em AV-convex-list} instead first produces two separate ranked lists by sorting the clips within each video according to the audio sampler and the visual sampler independently. Then the method computes for each clip the weighted average of its ranked position in the two lists according to a convex combination of the two positions. The top-$K$ clips according to this measure are finally retrieved. The method {\em AV-intersect-list} computes an intersection between the top-$m$ clips of the audio sampler and the top-$m$ clips of the video sampler. For each video, $m$ is progressively increased until the intersection yields exactly $K$ clips. In {\em AV-union-list} we form a set of $K$ clips by selecting $K'$-top clips according to the visual sampler (with hyperparameter $K'$ s.t. $K'<K$) and by adding to it a set of $K-K'$ different clips from the ranked list of the audio sampler. Finally, we also present results for {\em AV-joint-training}, where we simply average the audio-based score and the video-based score and then finetune the two networks with respect to this average.


\section{Experiments}

In this section we evaluate the proposed sampling procedure on the large-scale Sports1M and Kinetics datasets.

\subsection{Large-scale action recognition with SCSampler}

\subsubsection{Experimental Setup}
\label{sec:expsetup}

\noindent \textbf{Action Recognition Networks.}
    Our sampler can be used with any clip-based action classifier {\bf f}. We demonstrate the general applicability of our approach by evaluating it with six popular 3D CNNs for action recognition. Four of these models are pretrained networks publicly available~\cite{VMZ} and described in detail in~\cite{tran2017closer}: they are 18-layer instantiations of ResNet3D ({R3D}), Mixed Convolutional Network ({MC3}), and R(2+1)D, with this last network also in a 34-layer configuration. The other two models are our own implementation of I3D-RGB~\cite{carreira2017quo} and a ResNet3D of 152 layers leveraging depthwise convolutions (ir-CSN-152)~\cite{Tran19}. These networks are among the state-of-the-art on Kinetics and Sports1M. For training procedure, please refer to supplementary material. 

\noindent \textbf{Sampler configuration.}
In this subsection we present results achieved with the best configuration of our sampler architecture, based on the experimental study that we present in section~\ref{sec:abl_main}. The best configuration is a model that combines the saliency scores of an audio sampler and of a video sampler, using the strategy of {\em AV-union-list}. The video sampler is based on two ResNet-18 models trained on MD and RGB-R features, respectively, using the action classification loss ({\em AC}). The audio sampler is trained with the saliency ranking loss ({\em SAL-RANK}). Our sampler $s(.)$ is optimized with respect to the given clip classifier {\bf f}. Thus, we train a separate clip sampler for each of the 6 architectures in this evaluation. All results are based on sampling $K=10$ clips from the video, since this is the best  hyper-parameter value according to our experiments (see analysis in supplementary material). 

\noindent \textbf{Baselines.}
We compare the action recognition accuracy achieved with our sampler, against three baseline strategies to select $K=10$ clips from the video:  {\em Random} chooses clips at random, {\em Uniform} selects clips uniformly spaced out, while {\em Empirical} samples clips from the discrete empirical distribution (i.e., a histogram) of the top $K=10$ Oracle clip locations over the entire training set (the histogram is computed by linearly remapping the temporal extent of each video to be in the interval $[0,1]$). 
Finally, we also include video classification accuracy obtained with {\em Dense} which performs ``dense'' evaluation by averaging the clip-level predictions over all non-overlapping clips in the video.

\begin{table*}[]
\captionsetup{font=small}
\centering
\footnotesize
\vspace{7pt}
    \begin{minipage}{\textwidth}
        \centering
\begin{tabular}{@{}lccccccc@{}}
\toprule
\multicolumn{1}{c|}{Classifier} & \multicolumn{2}{c}{SCSampler $\mathcal{S}$ ($K$ clips)} & \multicolumn{2}{c}{Random / Uniform / Empirical ($K$ clips)} & \multicolumn{2}{c}{Dense ({\em all} clips)} & Oracle $\mathcal{O}$ ($K$ clips) \\
 & accuracy (\%) & \multicolumn{1}{c|}{runtime (day)} & accuracy (\%) & \multicolumn{1}{c|}{runtime (day)} & accuracy (\%) & \multicolumn{1}{c|}{runtime (days)} & accuracy (\%) \\ \midrule
\multicolumn{1}{c|}{MC3-18} & 72.8 & \multicolumn{1}{c|}{0.8} & 64.5 / 64.8 / 65.3
& \multicolumn{1}{c|}{0.4} & 66.6 & \multicolumn{1}{c|}{12.9} & 85.1 \\
\multicolumn{1}{c|}{R(2+1)D-18} & 73.9 & \multicolumn{1}{c|}{0.8} & 63.0 / 63.2 / 63.9 & \multicolumn{1}{c|}{0.4} & 68.7 & \multicolumn{1}{c|}{13.1} & 87.0 \\
\multicolumn{1}{c|}{R3D-18} & 70.2 & \multicolumn{1}{c|}{0.8} & 59.8 / 59.9 / 60.3 & \multicolumn{1}{c|}{0.4} & 65.6 & \multicolumn{1}{c|}{13.3} & 85.0 \\ \midrule
\multicolumn{1}{c|}{R(2+1)D-34} & 78.0 & \multicolumn{1}{c|}{0.9} & 71.2 / 71.5 / 72.0 & \multicolumn{1}{c|}{0.6} & 70.9 & \multicolumn{1}{c|}{14.2} & 88.4 \\
\multicolumn{1}{c|}{ir-CSN-152} & 84.0 & \multicolumn{1}{c|}{0.9} & 75.3 / 75.8 / 76.2 & \multicolumn{1}{c|}{0.5} & 77.0 & \multicolumn{1}{c|}{14.0} & 92.6 \\ \bottomrule
\end{tabular}

            \vspace{-6pt}
            \caption{\small Video-level classification on Sports1M~\cite{Sports1M} using $K$ clips selected by our SCSampler, chosen at ``Random'' or with ``Uniform'' spacing, by sampling clips according to the ``Empirical'' distribution computed on the training set, as well as  ``Dense'' evaluation on all clips. Oracle uses the {\em true label} of the test video to select clips. Runtime is the total time for evaluation over the entire test set. SCSampler delivers large gains over Dense, Random, Uniform and Empirical while keeping inference efficient. For ir-CSN-152, SCSampler yields a gain of $7.0\%$ over the already state-of-the-art accuracy of $77.0\%$ achieved by Dense.}
            \label{tab_sports}
            \vspace{5pt}
    \end{minipage}
    \begin{minipage}{\textwidth}
        \centering
\begin{tabular}{@{}cccccccc@{}}
\toprule
\multicolumn{1}{c|}{Classifier} & \multicolumn{2}{c}{SCSampler $\mathcal{S}$ ($K$ clips)} & \multicolumn{2}{c}{Random / Uniform / Empirical ($K$ clips)} & \multicolumn{2}{c}{Dense ({\em all} clips)} & Oracle $\mathcal{O}$ ($K$ clips) \\
 & accuracy (\%) & \multicolumn{1}{c|}{runtime (hr)} & accuracy (\%) & \multicolumn{1}{c|}{runtime (hr)} & accuracy (\%) & \multicolumn{1}{c|}{runtime (hr)} & accuracy (\%) \\ \midrule
\multicolumn{1}{c|}{MC3-18} & 67.0 & \multicolumn{1}{c|}{1.5} & 63.0 / 63.4 / 63.6 & \multicolumn{1}{c|}{1.3} & 65.1 & \multicolumn{1}{c|}{2.3} & 82.0 \\
\multicolumn{1}{c|}{R(2+1)D-18} & 70.9 & \multicolumn{1}{c|}{1.6} & 65.9 / 66.2 / 66.3 & \multicolumn{1}{c|}{1.4} & 68.0 & \multicolumn{1}{c|}{2.4} & 85.4 \\
\multicolumn{1}{c|}{R3D-18} & 67.3 & \multicolumn{1}{c|}{1.6} & 63.6 / 63.8 / 64.0 & \multicolumn{1}{c|}{1.3} & 65.2 & \multicolumn{1}{c|}{2.4} & 83.0 \\ \midrule
\multicolumn{1}{c|}{R(2+1D)-34*} & 76.7 & \multicolumn{1}{c|}{1.6} & 73.8 / 74.0 / 74.1 & \multicolumn{1}{c|}{1.5} & 74.1 & \multicolumn{1}{c|}{3.1} & 82.9 \\
\multicolumn{1}{c|}{I3D-RGB**} & 75.1 & \multicolumn{1}{c|}{1.5} & 71.9 / 71.8 / 71.9 & \multicolumn{1}{c|}{1.3} & 72.8 & \multicolumn{1}{c|}{2.9} & 81.2 \\
\multicolumn{1}{c|}{ir-CSN-152*} & 80.2 & \multicolumn{1}{c|}{1.6} & 77.8 / 78.5 / 79.2 & \multicolumn{1}{c|}{1.5} & 78.8 & \multicolumn{1}{c|}{3.0} & 89.0 \\ \bottomrule
\end{tabular}
            \vspace{-6pt}
            \caption{\small Video-level classification on Kinetics~\cite{Kinetics} using $K$ clips selected using our SCSampler, chosen at ``Random'' or with ``Uniform'' spacing, by sampling clips according to the ``Empirical'' distribution computed on the training set, as well as ``Dense'' evaluation on all clips. Even though Kinetics videos are short (10 seconds) our sampling procedure provides consistent accuracy gains for all 6 networks, compared to Random and Uniform clip selection or even Dense evaluation. Models marked with "*" are pretrained on Sports1M, and models with "**" are pretrained as 2D CNNs on ImageNet and then 3D-inflated~\cite{carreira2017quo}.}
            \label{tab:ls_kinetics}
            \vspace{5pt}
    \end{minipage}
    
    \begin{minipage}{\textwidth}
       \centering
\begin{tabular}{@{}c|ccccc|c@{}}
\toprule
 & \multicolumn{5}{c|}{Clip Selector} &  \\ \cmidrule(lr){2-6}
Test Set & \begin{tabular}[c]{@{}c@{}}SCSampler\\ Tr: MC3-18 on Kinetics\end{tabular} & \begin{tabular}[c]{@{}c@{}}SCSampler\\ Tr: MC3-18 on Sports1M\end{tabular} & \begin{tabular}[c]{@{}c@{}}SCSampler\\ Tr: R(2+1)D on Kinetics\end{tabular} & \begin{tabular}[c]{@{}c@{}}SCSampler\\ Tr: R(2+1)D on Sports1M\end{tabular} & Rand. / Unif. & Dense \\ \midrule
Kinetics & 67.0 & 65.0 & 65.9 & 65.0 & 63.1 / 62.3 & 65.1 \\
Sports1M & 69.2 & 72.8 & 68.5 & 72.1 & 64.6 / 64.8 & 66.6 \\ \bottomrule
\end{tabular}
            \vspace{-6pt}
            \caption{\small Cross-dataset and cross-classifier performance. Numbers report MC3-18 video-level accuracy on the validation set of Kinetics (first row) and test set of Sports1M (second row). SCSampler outperforms Uniform even when optimized for a different classifier (R(2+1)D) and a different dataset (e.g., 68.5\% vs 64.8\% for Sports1M).\vspace{-.5cm}}
            \label{tab:ls_crossDS}
    \end{minipage}

\end{table*}

\subsubsection{Evaluation on Sports1M}

Our approach is designed to operate on long, real-world videos where it is neither feasible
nor beneficial to evaluate every single clip. For these reasons, we choose the Sports1M dataset~\cite{Sports1M} as a suitable benchmark since its average video length is 5 minutes and 36 seconds, and some of its videos exceed 1 hour. We use the official training/test split. We do not trim the test videos and instead seek the top $K=10$ clips according to our sampler in each video. We stress that our sampling strategy is applied to test videos only. 
The training videos in Sports1M are also untrimmed. As training on all training clips would be unfeasible, we use the training procedure described in~\cite{tran2017closer} which consists in selecting from each training video 10 random 2-second segments, from which training clips are formed.
We reserve to future work the investigation of whether our sampling can be extended to sample {\em training} clips from the full videos.

We present the results in Table~\ref{tab_sports}, which includes for each method the video-level classification accuracy as well as the cumulative runtime (in days) to run the inference on the complete test set using 32 NVIDIA P100 GPUs (this includes the time needed for sampling as well as clip-level action classification). The most direct baselines for our evaluation are {\em Random}, {\em Uniform} and {\em Empirical} which use the same number of clips ($K$) in each video as SCSampler.
It can be seen that compared to these baselines, SCSampler delivers a substantial accuracy gain for all action models, with improvements ranging from 6.0\% for R(2+1)D-34 to 9.9\% for R(2+1)D-18 with respect to Empirical, which does only marginally better than Random and Uniform. 

Our approach does also better than ``Dense'' prediction, which averages the action classification predictions over {\em all} non-overlapping clips. To the best of our knowledge the accuracy of 77.0\% achieved by ir-CSN-152 using Dense evaluation is currently the best published result on this benchmark. SCSampler provides an additional gain of 7.0\% over this state-of-the-art model, pushing the accuracy to 84.0\%. We note that when using ir-CSN-152, Dense requires 14 days whereas SCSampler achieves better accuracy and requires only 0.65 days to run inference on the Sports1M test set. Finally, we report also the performance of the ``Oracle'' ${\mathcal O}$, which selects the $K$ clips that yield the highest classification score for the {\em true class} of the test video. This is an impractical model but it gives us an informative upper bound on the accuracy achievable with an ideal sampler.

Fig.~\ref{fig:histogram}~(left) shows the histogram of the clip temporal locations using $K=10$ samples per video for the test set of Sports1M (after remapping the temporal extent of each video to $[0,1]$). Oracle and SCSampler produce similar distributions of clip locations, with the first section and especially the last section of videos receiving many more samples. It can be noted that Empirical shows a different sample distribution compared to Oracle. This is due to the fact that it computes the histogram from the training set which in this case appears to have different statistics from the test set. 

\begin{figure}[t]
\captionsetup{font=small}
\small
\includegraphics[width=\columnwidth]{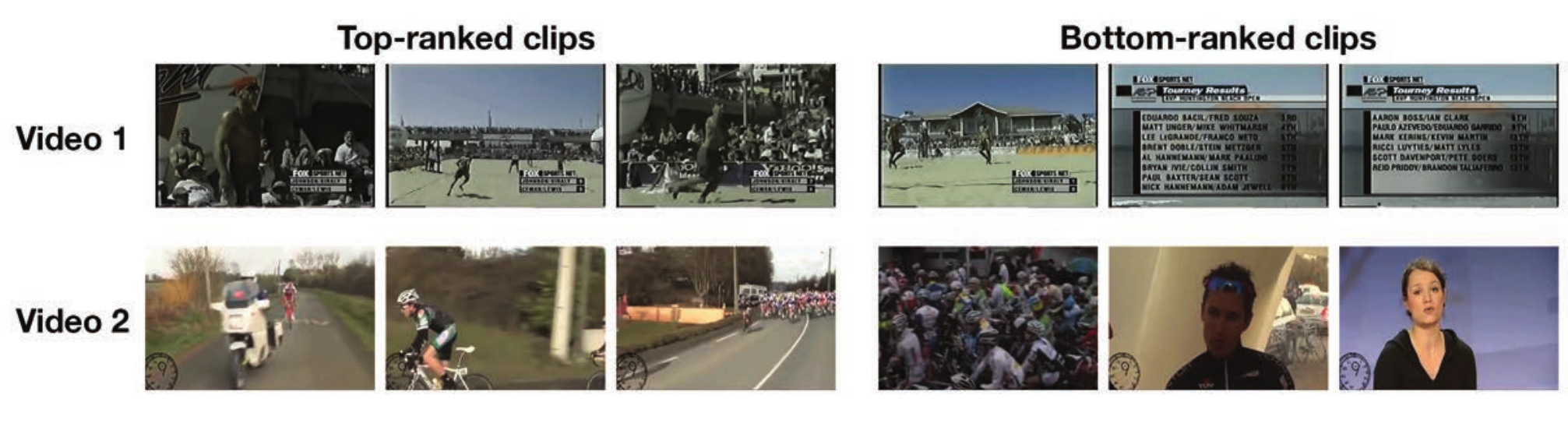}
\vspace{-10pt}
\caption{\small Top-ranked and bottom-ranked clips by SCSampler for two test videos from Sports1M. Top-ranked clips often show the sports in action, while bottom-ranked clips tend to be TV-interviews or static segments with scoreboard. Clips are shown as thumbnails. To see the videos please visit \textbf{http://scsampler.ai}.}
\label{fig:visualizations}
\vspace{-8pt}
\end{figure}

Thumbnails of top-ranked and bottom-ranked clips for two test videos are shown in Fig.~\ref{fig:visualizations}.

\subsubsection{Evaluation on Kinetics}
We further evaluate SCSampler on the Kinetics~\cite{Kinetics} dataset. Kinetics is a large-scale benchmark for action recognition containing 400 classes and 280K videos (240K for training and 40K for testing), each about 10 seconds long. The results are reported in Table~\ref{tab:ls_kinetics}. Kinetics videos are short and thus in principle the recognition model should not benefit from a clip-sampling scheme such as ours. Nevertheless, we see that for all architectures SCSampler provides accuracy gains over Random/Uniform/Empirical selection and Dense evaluation, although the improvements are understandably less substantial than in the case of Sports1M. To the best of our knowledge, the accuracy of 80.2\% achieved by ir-CSN-152 with our SCSampler is the best reported result so far on this benchmark.

Note that~\cite{Tran19} reports an accuracy of 79.0\% using Uniform (instead of the 78.5\% we list in Table~\ref{tab:ls_kinetics}, row 6) but this accuracy is achieved by applying the clip classifier spatially in a fully-convolutional fashion on frames of size 256x256, whereas here we use a single center spatial crop of size 224x224 for all our experiments. Sliding the clip classifier spatially in a fully-convolutional fashion (as in~\cite{Tran19}) raises the accuracy of SCSampler to 81.1\%.

\begin{figure}[t]
\includegraphics[width=\columnwidth]{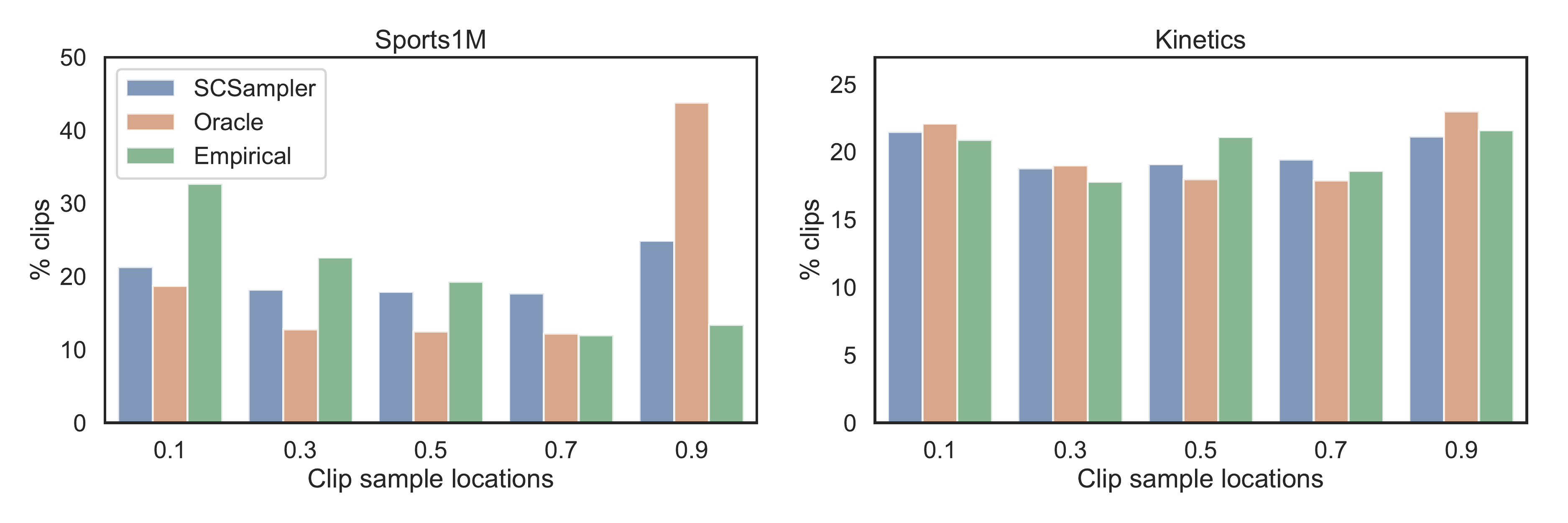}
\vspace{-15pt}
\caption{\small{Histogram of clip-sample locations on the test set of Sports1M (left) and validation set of Kinetics (right). The distribution of SCSampler matches fairly closely that of the Oracle.
}}
\label{fig:histogram}
\vspace{-8pt}

\end{figure}

Fig.~\ref{fig:histogram}~(right) shows the histogram of clip temporal locations on the validation set of Kinetics. Compared to Sports1M, the Oracle and SCSampler distributions here is much more uniform.

\subsubsection{Unseen Action Classifiers and Novel Classes}

While our SCSampler has low computational cost, it adds the procedural overhead of having to train a specialized clip selector for each classifier and each dataset. Here we evaluate the possibility of reusing a sampler $s(.)$ that was optimized for a classifier ${\bf f}$ on a dataset $\mathcal{D}$, for a new classifier  ${\bf f}'$ on a dataset $\mathcal{D}'$ that contains action classes different from those seen in $\mathcal{D}$. In Table~\ref{tab:ls_crossDS}, we present cross-dataset performance of an SCSampler trained on Kinetics but then used to select clips on Sports1M (and vice-versa). We also report cross-classifier performance obtained by optimizing SCSampler with pseudo-ground truth labels (see section~\ref{sec:salrankloss}) generated by R(2+1)D-18 but then used for video-level prediction with action classifier MC3-18. On the Kinetics validation set, using an SCSampler that was trained using the same action classifier (MC3) but a different dataset (Sports1M) causes a drop of about 2\% (65.0\% vs 67.0\%) while training using a different action classifier (R(2+1)D) to generate pseudo-ground truth labels on the the same dataset (Kinetics) causes a degradation of 1.1\% (65.9\% vs 67.0\%). The evaluation on Sports1M shows a similar trend, where cross-dataset accuracy (69.2\%) is lower than cross-classifier accuracy (72.1\%). 
Even in the extreme setting of cross-dataset {\em and} cross-classifier, the accuracies achieved with SCSampler are still better than those obtained with Random or Uniform selection. Finally, we note that samplers trained using the $AC$ loss (section~\ref{sec:acloss}) do not require pseudo-labels and thus are independent of the action classifier by design.

\subsection{Evaluating Design Choices for SCSampler}\label{sec:abl_main}

In this subsection we evaluate the different choices in the design of SCSampler. Given the many configurations to assess, we make this  study more computationally feasible by restricting the evaluation to a subset of Sports1M, which we name {\em miniSports}. The dataset is formed by randomly choosing for each class 280 videos from the training set and 69 videos from the test set. This gives us a class-balanced set of 136,360 training videos and 33,603 test videos. All videos are shortened to the same length of 2.75 minutes. For our assessment, we restrict our choice of action classifier to MC3-18, which we retrain on our training set of miniSports. We assess the SCSampler design choices in terms of how they affect the video-level accuracy of MC3-18 on the test set of miniSports, since our aim is to find the best configuration for video classification.

\subsubsection{Learning objective}\label{sec:abl_loss}

We begin by studying the effect of the loss function used for training SCSampler, by considering the two loss variants described in section~\ref{subs:loss_choice}. For this evaluation, we assess separately the visual sampler and the audio sampler. The video sampler is based on two ResNet-18 networks with MD and RGB-R features, respectively. These 2 networks are pretrained on ImageNet and then finetuned on the training set of miniSport for each of the three different SCSampler loss functions. The audio sampler is our VGG network pretrained for classification on AudioSet~\cite{audioset} and then finetuned on the training set of miniSports. The MC3-18 video classification accuracy is 73.1\% when the visual sampler is trained with the Action Classification (AC) loss whereas it is 64.8\% when it is trained with the Saliency Ranking (SAL-RANK) loss. Conversely, we found that the audio sampler is slightly more effective when trained with the SAL-RANK loss as opposed to the AC loss (video-level accuracy is 67.8\% with SAL-RANK and 66.4\% with AC). A possible explanation for this difference in results is that the AC loss defines a more challenging problem to address (action classification vs binary ranking) but provides more supervision (multiclass vs binary labels). The model using compressed video features is a stronger model that can benefit from the AC supervision and do well on this task (as already shown in~\cite{wu2018coviar}) but the weaker audio model does better when trained on the simpler SAL-RANK problem.

\subsubsection{Sampler architecture and features}\label{sec:abl_archfeat}

In this subsection we assess different architectures and features for the sampler. For the visual sampler, we use the AC loss and
consider two different lightweight architectures: ResNet-18 and ShuffleNet26. Each architecture is trained on each of the 3 types of video-compression features described in section~\ref{subsubvisualsampler}: IF, MD and RGB-R. We also assess performance of combination of these three features by averaging the scores of classifiers based on individual features. The results are reported in Table~\ref{tab:visual_features}. We can observe that given the same type of input features, ResNet-18 provides much higher accuracy than ShuffeNet-26 at a runtime that is only marginally higher. It can be noticed that MD and RGB-R features seem to be quite complementary: for ResNet-18, MD+RGB-R yields an accuracy of 73.1\% whereas these individual features alone achieve an accuracy of only 68.0\% and 63.5\%. However, adding IF features to MD+RGB-R provides a modest gain in accuracy (74.9 vs 73.1) but impacts noticeably the runtime. Considering these tradeoffs, we adopt ResNet-18 trained on MD+RGB-R as our visual sampler on all subsequent experiments.

We perform a similar ablation study for the audio sampler. Given our VGG audio network pretrained for classification on AudioSet, we train it on miniSport using the following two options: finetuning the entire VGG model  vs training a single FC layer on several VGG activations. Finetuning the audio sampler yields the best classification accuracy (see detailed results in supplementary material).

\begin{table}[t]
\captionsetup{font=small}
\footnotesize
\centering
\begin{tabular}{@{}cccc@{}}
\toprule
\begin{tabular}[c]{@{}c@{}}SCSampler\\ features \end{tabular}  & \begin{tabular}[c]{@{}c@{}}SCSampler\\ architecture \end{tabular} & accuracy (\%) & \begin{tabular}[c]{@{}c@{}}runtime\\ (min)\end{tabular} \\ \midrule
MD & \multicolumn{1}{c|}{ResNet-18} & 63.5 & 19.8 \\
RGB-R & \multicolumn{1}{c|}{ResNet-18} & 68.0 & 20.4 \\
MD + RGB-R & \multicolumn{1}{c|}{ResNet-18} & 73.1 & 20.9 \\
IF+MD+RGB-R & \multicolumn{1}{c|}{ResNet-18} & 74.9 & 27.3 \\
MD + RGB-R & \multicolumn{1}{c|}{ShuffleNet-26} & 67.9 & 19.1 \\
IF+MD+RGB-R & \multicolumn{1}{c|}{ShuffleNet-26} & 69.9 & 23.8 \\ \midrule
\end{tabular}
\vspace{-6pt}
\caption{\small Varying the visual sampler architecture (ResNet-18 vs ShuffleNet-26) and the input compressed channel (IF, MD, or RGB-R). Performance is measured as video-level accuracy (\%) achieved by MC3-18 on the miniSports  test set with $K=10$ sampled clips. Runtime is on the full test set using 32 GPUs.}
\label{tab:visual_features}
\vspace{-7pt}

\end{table}

\subsubsection{Combining audio and visual saliency}

In this subsection we assess the impact of our different schemes for combining audio-based and video-based saliency scores (see~\ref{subsubcombined}). For this we use the best configurations of our visual and audio sampler (described in~\ref{sec:expsetup}). Table~\ref{tab:combined_features} shows the video-level action recognition accuracy achieved for the different combination strategies. 

Perhaps surprisingly, the best results are achieved with {\em AV-union-list}, which is the rather na\"ive solution of taking $K'$ clips based on the video sampler and $K-K'$ different clips based on the audio sampler ($K'=8$ is the best value when $K=10$). The more sophisticated approach of joint training {\em AV-joint-training} performs nearly on-par with it. 
Overall, it is clear that the visual sampler is a better clip selector than the audio sampler. But considering the small cost of audio-based sampling, the accuracy gain provided by {\em AV-union-list} over visual only (76.0 vs 73.1) warrants the use of this combination.

\begin{table}[t]
\captionsetup{font=small}
\footnotesize
\centering
\begin{tabular}{@{}ccc@{}}
\toprule
\begin{tabular}[c]{@{}c@{}}SCSampler \\Audio-Video Combination\end{tabular} & accuracy (\%) & runtime (min) \\ \midrule
\begin{tabular}[c]{@{}c@{}}{AV-convex-list} ($\alpha = 0.8$)\\ \end{tabular} & 73.8 & 23.4 \\
\begin{tabular}[c]{@{}c@{}}{AV-convex-score} ($\alpha = 0.9$)\\ \end{tabular} & 67.9 & 23.4 \\
\begin{tabular}[c]{@{}c@{}}{AV-union-list} ($K'=8$)\end{tabular} & 76.0 & 23.4 \\
{AV-intersect-list} & 74.0 & 23.4 \\
{AV-joint-training} & 75.5 & 23.4 \\ \midrule
Visual SCSampler only & 73.1 & 20.9 \\
Audio SCSampler only & 67.8 & 22.0 \\ \midrule
Random & 59.5 & 15.1 \\
Uniform & 59.9 & 15.1 \\
Dense & 61.6 & 2293.5 (38.5 hrs) \\ \bottomrule
\end{tabular}
\vspace{-6pt}
\caption{\small Different schemes of combining audio and video saliency. Performance is measured as MC3-18 video classification accuracy (\%) on the test set of miniSports with $K=10$ sampled clips.}
\label{tab:combined_features}
\vspace{-6pt}

\end{table}

\section{Discussion}

We presented a very simple scheme to boost both the accuracy and the speed of clip-based action classifiers. It leverages a lightweight clip-sampling model to select a small subset of clips for analysis. Experiments show that, despite its simplicity, our  clip-sampler yields large accuracy gains and big speedups for 6 different strong action recognizers, and it retains strong performance even when used on novel classes. 
Future work will investigate strategies for optimal sample-set selection, by taking into account clip redundancies. It would be interesting to extend our sampling scheme to models that employ more sophisticated aggregations than simple averaging, e.g., those that use a set of contiguous clips to capture long-range temporal structure. SCSampler scores for the test videos of Kinetics and Sports1M are available for download at \textbf{http://scsampler.ai}.

{\small
\section*{Acknowledgments}
We would like to thank Zheng Shou and Chao-Yuan Wu for providing help with reading and processing of compressed video.
}

{\small
\bibliographystyle{ieee_fullname}
\bibliography{main.bib}

\begin{thebibliography}{10}\itemsep=-1pt

\bibitem{ActionSearch}
Humam Alwassel, Fabian Caba~Heilbron, and Bernard Ghanem.
\newblock Action search: Spotting actions in videos and its application to
  temporal action localization.
\newblock In Vittorio Ferrari, Martial Hebert, Cristian Sminchisescu, and Yair
  Weiss, editors, {\em Computer Vision -- ECCV 2018}, pages 253--269, Cham,
  2018. Springer International Publishing.

\bibitem{arandjelovic2017look}
Relja Arandjelovic and Andrew Zisserman.
\newblock Look, listen and learn.
\newblock In {\em {IEEE} International Conference on Computer Vision, {ICCV}
  2017, Venice, Italy, October 22-29, 2017}, pages 609--617, 2017.

\bibitem{arandjelovic2017objects}
Relja Arandjelovic and Andrew Zisserman.
\newblock Objects that sound.
\newblock In {\em Computer Vision - {ECCV} 2018 - 15th European Conference,
  Munich, Germany, September 8-14, 2018, Proceedings, Part {I}}, pages
  451--466, 2018.

\bibitem{aytar2016soundnet}
Yusuf Aytar, Carl Vondrick, and Antonio Torralba.
\newblock Soundnet: Learning sound representations from unlabeled video.
\newblock In {\em Advances in Neural Information Processing Systems 29: Annual
  Conference on Neural Information Processing Systems 2016, December 5-10,
  2016, Barcelona, Spain}, pages 892--900, 2016.

\bibitem{BuchEtAl:CVPR17}
Shyamal Buch, Victor Escorcia, Chuanqi Shen, Bernard Ghanem, and Juan~Carlos
  Niebles.
\newblock {SST:} single-stream temporal action proposals.
\newblock In {\em 2017 {IEEE} Conference on Computer Vision and Pattern
  Recognition, {CVPR} 2017, Honolulu, HI, USA, July 21-26, 2017}, pages
  6373--6382, 2017.

\bibitem{carreira2017quo}
Jo{\~{a}}o Carreira and Andrew Zisserman.
\newblock Quo vadis, action recognition? {A} new model and the kinetics
  dataset.
\newblock In {\em 2017 {IEEE} Conference on Computer Vision and Pattern
  Recognition, {CVPR} 2017, Honolulu, HI, USA, July 21-26, 2017}, pages
  4724--4733, 2017.

\bibitem{chung2016out}
Joon~Son Chung and Andrew Zisserman.
\newblock Out of time: Automated lip sync in the wild.
\newblock In {\em Computer Vision - {ACCV} 2016 Workshops - {ACCV} 2016
  International Workshops, Taipei, Taiwan, November 20-24, 2016, Revised
  Selected Papers, Part {II}}, pages 251--263, 2016.

\bibitem{EscorciaEtal:ECCV16}
Victor Escorcia, Fabian~Caba Heilbron, Juan~Carlos Niebles, and Bernard Ghanem.
\newblock Daps: Deep action proposals for action understanding.
\newblock In {\em Computer Vision - {ECCV} 2016 - 14th European Conference,
  Amsterdam, The Netherlands, October 11-14, 2016, Proceedings, Part {III}},
  pages 768--784, 2016.

\bibitem{VMZ}
Facebook.
\newblock Video model zoo.
\newblock \url{https://github.com/facebookresearch/VMZ}, 2018.

\bibitem{fan18ijcai}
Hehe Fan, Zhongwen Xu, Linchao Zhu, Chenggang Yan, Jianjun Ge, and Yi Yang.
\newblock Watching a small portion could be as good as watching all: Towards
  efficient video classification.
\newblock In {\em International Joint Conference on Artificial Intelligence,
  {IJCAI} 2018, Stockholm, Sweden, July 13-19, 2018}, pages 705--711, 2018.

\bibitem{SlowFast}
Christoph Feichtenhofer, Haoqi Fan, Jitendra Malik, and Kaiming He.
\newblock Slowfast networks for video recognition.
\newblock {\em CoRR}, abs/1812.03982, 2018.

\bibitem{feichtenhofer2016spatiotemporal}
Christoph Feichtenhofer, Axel Pinz, and Richard Wildes.
\newblock Spatiotemporal residual networks for video action recognition.
\newblock In {\em Advances in neural information processing systems}, pages
  3468--3476, 2016.

\bibitem{GaidonEtAl:TPAMI2013}
Adrien Gaidon, Za{\"{\i}}d Harchaoui, and Cordelia Schmid.
\newblock Temporal localization of actions with actoms.
\newblock {\em {IEEE} Trans. Pattern Anal. Mach. Intell.}, 35(11):2782--2795,
  2013.

\bibitem{GaoEtAl:ICCV17}
Jiyang Gao, Zhenheng Yang, Chen Sun, Kan Chen, and Ram Nevatia.
\newblock {TURN} {TAP:} temporal unit regression network for temporal action
  proposals.
\newblock In {\em {IEEE} International Conference on Computer Vision, {ICCV}
  2017, Venice, Italy, October 22-29, 2017}, pages 3648--3656, 2017.

\bibitem{Gao18ECCV}
Ruohan Gao, Rog{\'{e}}rio~Schmidt Feris, and Kristen Grauman.
\newblock Learning to separate object sounds by watching unlabeled video.
\newblock In {\em 2018 {IEEE} Conference on Computer Vision and Pattern
  Recognition Workshops, {CVPR} Workshops 2018, Salt Lake City, UT, USA, June
  18-22, 2018}, pages 2496--2499, 2018.

\bibitem{audioset}
Jort~F. Gemmeke, Daniel P.~W. Ellis, Dylan Freedman, Aren Jansen, Wade
  Lawrence, R.~Channing Moore, Manoj Plakal, and Marvin Ritter.
\newblock Audio set: An ontology and human-labeled dataset for audio events.
\newblock In {\em Proc. IEEE ICASSP 2017}, New Orleans, LA, 2017.

\bibitem{Ghirdar:CVPR17}
Rohit Girdhar, Deva Ramanan, Abhinav Gupta, Josef Sivic, and Bryan Russell.
\newblock Actionvlad: Learning spatio-temporal aggregation for action
  classification.
\newblock In {\em 2017 {IEEE} Conference on Computer Vision and Pattern
  Recognition, {CVPR} 2017, Honolulu, HI, USA, July 21-26, 2017}, pages
  3165--3174, 2017.

\bibitem{GongEtAl:NIPS16}
Boqing Gong, Wei{-}Lun Chao, Kristen Grauman, and Fei Sha.
\newblock Diverse sequential subset selection for supervised video
  summarization.
\newblock In {\em Advances in Neural Information Processing Systems 27: Annual
  Conference on Neural Information Processing Systems 2014, December 8-13 2014,
  Montreal, Quebec, Canada}, pages 2069--2077, 2014.

\bibitem{GygliEtAl:CVPR15}
Michael Gygli, Helmut Grabner, and Luc J.~Van Gool.
\newblock Video summarization by learning submodular mixtures of objectives.
\newblock In {\em {IEEE} Conference on Computer Vision and Pattern Recognition,
  {CVPR} 2015, Boston, MA, USA, June 7-12, 2015}, pages 3090--3098, 2015.

\bibitem{he2016residual}
Kaiming He, Xiangyu Zhang, Shaoqing Ren, and Jian Sun.
\newblock Deep residual learning for image recognition.
\newblock In {\em 2016 {IEEE} Conference on Computer Vision and Pattern
  Recognition, {CVPR} 2016, Las Vegas, NV, USA, June 27-30, 2016}, pages
  770--778, 2016.

\bibitem{HeilbronEtAl:CVPR16}
Fabian~Caba Heilbron, Juan~Carlos Niebles, and Bernard Ghanem.
\newblock Fast temporal activity proposals for efficient detection of human
  actions in untrimmed videos.
\newblock In {\em 2016 {IEEE} Conference on Computer Vision and Pattern
  Recognition, {CVPR} 2016, Las Vegas, NV, USA, June 27-30, 2016}, pages
  1914--1923, 2016.

\bibitem{Mihir:CVPR14}
Mihir Jain, Jan~C. van Gemert, Herv{\'{e}} J{\'{e}}gou, Patrick Bouthemy, and
  Cees G.~M. Snoek.
\newblock Action localization with tubelets from motion.
\newblock In {\em 2014 {IEEE} Conference on Computer Vision and Pattern
  Recognition, {CVPR} 2014, Columbus, OH, USA, June 23-28, 2014}, pages
  740--747, 2014.

\bibitem{R1:MR1}
Vadim Kantorov and Ivan Laptev.
\newblock Efficient feature extraction, encoding, and classification for action
  recognition.
\newblock In {\em 2014 {IEEE} Conference on Computer Vision and Pattern
  Recognition, {CVPR} 2014, Columbus, OH, USA, June 23-28, 2014}, pages
  2593--2600, 2014.

\bibitem{Sports1M}
Andrej Karpathy, George Toderici, Sanketh Shetty, Thomas Leung, Rahul
  Sukthankar, and Fei{-}Fei Li.
\newblock Large-scale video classification with convolutional neural networks.
\newblock In {\em 2014 {IEEE} Conference on Computer Vision and Pattern
  Recognition, {CVPR} 2014, Columbus, OH, USA, June 23-28, 2014}, pages
  1725--1732, 2014.

\bibitem{Kinetics}
Will Kay, Jo{\~{a}}o Carreira, Karen Simonyan, Brian Zhang, Chloe Hillier,
  Sudheendra Vijayanarasimhan, Fabio Viola, Tim Green, Trevor Back, Paul
  Natsev, Mustafa Suleyman, and Andrew Zisserman.
\newblock The kinetics human action video dataset.
\newblock {\em CoRR}, abs/1705.06950, 2017.

\bibitem{korbar2018cooperative}
Bruno Korbar, Du Tran, and Lorenzo Torresani.
\newblock Cooperative learning of audio and video models from self-supervised
  synchronization.
\newblock In {\em Advances in Neural Information Processing Systems 31: Annual
  Conference on Neural Information Processing Systems 2018, NeurIPS 2018, 3-8
  December 2018, Montr{\'{e}}al, Canada.}, pages 7774--7785, 2018.

\bibitem{LinEtAl:CVPRW17}
Tianwei Lin, Xu Zhao, and Zheng Shou.
\newblock Temporal convolution based action proposal: Submission to activitynet
  2017.
\newblock {\em CoRR}, abs/1707.06750, 2017.

\bibitem{LinEtAl:ECCV18}
Tianwei Lin, Xu Zhao, Haisheng Su, Chongjing Wang, and Ming Yang.
\newblock {BSN:} boundary sensitive network for temporal action proposal
  generation.
\newblock In {\em Computer Vision - {ECCV} 2018 - 15th European Conference,
  Munich, Germany, September 8-14, 2018, Proceedings, Part {IV}}, pages 3--21,
  2018.

\bibitem{MahasseniEtAl:CVPR17}
Behrooz Mahasseni, Michael Lam, and Sinisa Todorovic.
\newblock Unsupervised video summarization with adversarial lstm networks.
\newblock In {\em The IEEE Conference on Computer Vision and Pattern
  Recognition (CVPR)}, July 2017.

\bibitem{MerlerEtAl:CVPRW18}
Michele Merler, Dhiraj Joshi, Khoi-Nguyen~C. Mac, Quoc-Bao Nguyen, Stephen
  Hammer, John Kent, Jinjun Xiong, Minh~N. Do, John~R. Smith, and Rogerio~S.
  Feris.
\newblock The excitement of sports: Automatic highlights using audio/visual
  cues.
\newblock In {\em The IEEE Conference on Computer Vision and Pattern
  Recognition (CVPR) Workshops}, June 2018.

\bibitem{MerlerEtAl:TMM18}
M. {Merler}, K.~C. {Mac}, D. {Joshi}, Q. {Nguyen}, S. {Hammer}, J. {Kent}, J.
  {Xiong}, M.~N. {Do}, J.~R. {Smith}, and R. {Feris}.
\newblock Automatic curation of sports highlights using multimodal excitement
  features.
\newblock {\em IEEE Transactions on Multimedia}, pages 1--1, 2018.

\bibitem{R1:MR3}
Antoine Miech, Ivan Laptev, and Josef Sivic.
\newblock Learnable pooling with context gating for video classification.
\newblock {\em CoRR}, abs/1706.06905, 2017.

\bibitem{NgEtAl:CVPR15}
Joe~Yue{-}Hei Ng, Matthew~J. Hausknecht, Sudheendra Vijayanarasimhan, Oriol
  Vinyals, Rajat Monga, and George Toderici.
\newblock Beyond short snippets: Deep networks for video classification.
\newblock In {\em {IEEE} Conference on Computer Vision and Pattern Recognition,
  {CVPR} 2015, Boston, MA, USA, June 7-12, 2015}, pages 4694--4702, 2015.

\bibitem{yue2015beyond}
Joe~Yue{-}Hei Ng, Matthew~J. Hausknecht, Sudheendra Vijayanarasimhan, Oriol
  Vinyals, Rajat Monga, and George Toderici.
\newblock Beyond short snippets: Deep networks for video classification.
\newblock In {\em {IEEE} Conference on Computer Vision and Pattern Recognition,
  {CVPR} 2015, Boston, MA, USA, June 7-12, 2015}, pages 4694--4702, 2015.

\bibitem{owens2018audio}
Andrew Owens and Alexei~A. Efros.
\newblock Audio-visual scene analysis with self-supervised multisensory
  features.
\newblock In {\em Computer Vision - {ECCV} 2018 - 15th European Conference,
  Munich, Germany, September 8-14, 2018, Proceedings, Part {VI}}, pages
  639--658, 2018.

\bibitem{PirsiavashEtAl:CVPR14}
Hamed Pirsiavash and Deva Ramanan.
\newblock Parsing videos of actions with segmental grammars.
\newblock In {\em 2014 {IEEE} Conference on Computer Vision and Pattern
  Recognition, {CVPR} 2014, Columbus, OH, USA, June 23-28, 2014}, pages
  612--619, 2014.

\bibitem{PotapovEtAl:ECCV14}
Danila Potapov, Matthijs Douze, Za{\"{\i}}d Harchaoui, and Cordelia Schmid.
\newblock Category-specific video summarization.
\newblock In {\em Computer Vision - {ECCV} 2014 - 13th European Conference,
  Zurich, Switzerland, September 6-12, 2014, Proceedings, Part {VI}}, pages
  540--555, 2014.

\bibitem{ILSVRC15}
Olga Russakovsky, Jia Deng, Hao Su, Jonathan Krause, Sanjeev Satheesh, Sean Ma,
  Zhiheng Huang, Andrej Karpathy, Aditya Khosla, Michael Bernstein,
  Alexander~C. Berg, and Li Fei-Fei.
\newblock {ImageNet Large Scale Visual Recognition Challenge}.
\newblock {\em International Journal of Computer Vision (IJCV)},
  115(3):211--252, 2015.

\bibitem{ShouEtAl:CVPR17}
Zheng Shou, Jonathan Chan, Alireza Zareian, Kazuyuki Miyazawa, and Shih{-}Fu
  Chang.
\newblock {CDC:} convolutional-de-convolutional networks for precise temporal
  action localization in untrimmed videos.
\newblock In {\em 2017 {IEEE} Conference on Computer Vision and Pattern
  Recognition, {CVPR} 2017, Honolulu, HI, USA, July 21-26, 2017}, pages
  1417--1426, 2017.

\bibitem{ShouEtAL:CVPR16}
Zheng Shou, Dongang Wang, and Shih{-}Fu Chang.
\newblock Temporal action localization in untrimmed videos via multi-stage
  cnns.
\newblock In {\em 2016 {IEEE} Conference on Computer Vision and Pattern
  Recognition, {CVPR} 2016, Las Vegas, NV, USA, June 27-30, 2016}, pages
  1049--1058, 2016.

\bibitem{TwoStreamAZ:NIPS14}
Karen Simonyan and Andrew Zisserman.
\newblock Two-stream convolutional networks for action recognition in videos.
\newblock In {\em Advances in Neural Information Processing Systems 27: Annual
  Conference on Neural Information Processing Systems 2014, December 8-13 2014,
  Montreal, Quebec, Canada}, pages 568--576, 2014.

\bibitem{vgg2014}
Karen Simonyan and Andrew Zisserman.
\newblock Very deep convolutional networks for large-scale image recognition.
\newblock {\em CoRR}, abs/1409.1556, 2014.

\bibitem{tran2015learning}
Du Tran, Lubomir Bourdev, Rob Fergus, Lorenzo Torresani, and Manohar Paluri.
\newblock Learning spatiotemporal features with 3d convolutional networks.
\newblock In {\em 2015 {IEEE} International Conference on Computer Vision,
  {ICCV} 2015, Santiago, Chile, December 7-13, 2015}, pages 4489--4497, 2015.

\bibitem{Tran19}
Du Tran, Heng Wang, Lorenzo Torresani, and Matt Feiszli.
\newblock Classification with channel-separated convolutional networks.
\newblock In {\em {IEEE} International Conference on Computer Vision, {ICCV}
  2019, Seoul, South Korea, October 26-November 2, 2019}, 2019.

\bibitem{tran2017closer}
Du Tran, Heng Wang, Lorenzo Torresani, Jamie Ray, Yann LeCun, and Manohar
  Paluri.
\newblock A closer look at spatiotemporal convolutions for action recognition.
\newblock In {\em 2018 {IEEE} Conference on Computer Vision and Pattern
  Recognition, {CVPR} 2018, Salt Lake City, UT, USA, June 18-22, 2018}, pages
  6450--6459, 2018.

\bibitem{VarolEtAl:TPAMI17}
G{\"{u}}l Varol, Ivan Laptev, and Cordelia Schmid.
\newblock Long-term temporal convolutions for action recognition.
\newblock {\em {IEEE} Trans. Pattern Anal. Mach. Intell.}, 40(6):1510--1517,
  2018.

\bibitem{WangCherian:ECCV18}
Jue Wang and Anoop Cherian.
\newblock Learning discriminative video representations using adversarial
  perturbations.
\newblock In {\em Computer Vision - {ECCV} 2018 - 15th European Conference,
  Munich, Germany, September 8-14, 2018, Proceedings, Part {IV}}, pages
  716--733, 2018.

\bibitem{WangEtaAl:IJCV16}
Limin Wang, Yu Qiao, and Xiaoou Tang.
\newblock Mofap: {A} multi-level representation for action recognition.
\newblock {\em International Journal of Computer Vision}, 119(3):254--271,
  2016.

\bibitem{WangEtAl_TSN:ECCV16}
Limin Wang, Yuanjun Xiong, Zhe Wang, Yu Qiao, Dahua Lin, Xiaoou Tang, and
  Luc~Van Gool.
\newblock Temporal segment networks: Towards good practices for deep action
  recognition.
\newblock In {\em Computer Vision - {ECCV} 2016 - 14th European Conference,
  Amsterdam, The Netherlands, October 11-14, 2016, Proceedings, Part {VIII}},
  pages 20--36, 2016.

\bibitem{WangEtAl:CVPR16}
Xiaolong Wang, Ali Farhadi, and Abhinav Gupta.
\newblock Actions {\textasciitilde} transformations.
\newblock In {\em 2016 {IEEE} Conference on Computer Vision and Pattern
  Recognition, {CVPR} 2016, Las Vegas, NV, USA, June 27-30, 2016}, pages
  2658--2667, 2016.

\bibitem{NonLocal}
Xiaolong Wang, Ross~B. Girshick, Abhinav Gupta, and Kaiming He.
\newblock Non-local neural networks.
\newblock In {\em 2018 {IEEE} Conference on Computer Vision and Pattern
  Recognition, {CVPR} 2018, Salt Lake City, UT, USA, June 18-22, 2018}, pages
  7794--7803, 2018.

\bibitem{WuEtAl:CVPR19}
Chao{-}Yuan Wu, Christoph Feichtenhofer, Haoqi Fan, Kaiming He, Philipp
  Kr{\"{a}}henb{\"{u}}hl, and Ross~B. Girshick.
\newblock Long-term feature banks for detailed video understanding.
\newblock {\em CoRR}, abs/1812.05038, 2018.

\bibitem{wu2018coviar}
Chao{-}Yuan Wu, Manzil Zaheer, Hexiang Hu, R. Manmatha, Alexander~J. Smola, and
  Philipp Kr{\"{a}}henb{\"{u}}hl.
\newblock Compressed video action recognition.
\newblock In {\em 2018 {IEEE} Conference on Computer Vision and Pattern
  Recognition, {CVPR} 2018, Salt Lake City, UT, USA, June 18-22, 2018}, pages
  6026--6035, 2018.

\bibitem{Wu_2019_CVPR}
Zuxuan Wu, Caiming Xiong, Chih-Yao Ma, Richard Socher, and Larry~S. Davis.
\newblock Adaframe: Adaptive frame selection for fast video recognition.
\newblock In {\em The IEEE Conference on Computer Vision and Pattern
  Recognition (CVPR)}, June 2019.

\bibitem{xu2017rc3d}
Huijuan Xu, Abir Das, and Kate Saenko.
\newblock {R-C3D:} region convolutional 3d network for temporal activity
  detection.
\newblock In {\em {IEEE} International Conference on Computer Vision, {ICCV}
  2017, Venice, Italy, October 22-29, 2017}, pages 5794--5803, 2017.

\bibitem{yeung2016end}
Serena Yeung, Olga Russakovsky, Greg Mori, and Li Fei-Fei.
\newblock End-to-end learning of action detection from frame glimpses in
  videos.
\newblock In {\em Proceedings of the IEEE Conference on Computer Vision and
  Pattern Recognition}, pages 2678--2687, 2016.

\bibitem{zhang2016cvpr}
Bowen Zhang, Limin Wang, Zhe Wang, Yu Qiao, and Hanli Wang.
\newblock Real-time action recognition with enhanced motion vector cnns.
\newblock In {\em 2016 {IEEE} Conference on Computer Vision and Pattern
  Recognition, {CVPR} 2016, Las Vegas, NV, USA, June 27-30, 2016}, pages
  2718--2726, 2016.

\bibitem{ZhangEtAl:CVPR16}
Ke Zhang, Wei{-}Lun Chao, Fei Sha, and Kristen Grauman.
\newblock Summary transfer: Exemplar-based subset selection for video
  summarization.
\newblock In {\em 2016 {IEEE} Conference on Computer Vision and Pattern
  Recognition, {CVPR} 2016, Las Vegas, NV, USA, June 27-30, 2016}, pages
  1059--1067, 2016.

\bibitem{ZhangEtAl:ECCV16}
Ke Zhang, Wei{-}Lun Chao, Fei Sha, and Kristen Grauman.
\newblock Video summarization with long short-term memory.
\newblock In {\em Computer Vision - {ECCV} 2016 - 14th European Conference,
  Amsterdam, The Netherlands, October 11-14, 2016, Proceedings, Part {VII}},
  pages 766--782, 2016.

\bibitem{zhang2018shufflenet}
Xiangyu Zhang, Xinyu Zhou, Mengxiao Lin, and Jian Sun.
\newblock Shufflenet: An extremely efficient convolutional neural network for
  mobile devices.
\newblock In {\em 2018 {IEEE} Conference on Computer Vision and Pattern
  Recognition, {CVPR} 2018, Salt Lake City, UT, USA, June 18-22, 2018}, pages
  6848--6856, 2018.

\bibitem{zhao2018sound}
Hang Zhao, Chuang Gan, Andrew Rouditchenko, Carl Vondrick, Josh~H. McDermott,
  and Antonio Torralba.
\newblock The sound of pixels.
\newblock In {\em Computer Vision - {ECCV} 2018 - 15th European Conference,
  Munich, Germany, September 8-14, 2018, Proceedings, Part {I}}, pages
  587--604, 2018.

\bibitem{ZhaoEtAl:ICCV17}
Yue Zhao, Yuanjun Xiong, Limin Wang, Zhirong Wu, Xiaoou Tang, and Dahua Lin.
\newblock Temporal action detection with structured segment networks.
\newblock In {\em {IEEE} International Conference on Computer Vision, {ICCV}
  2017, Venice, Italy, October 22-29, 2017}, pages 2933--2942, 2017.

\bibitem{ZhouEtAl:ECCV18}
Bolei Zhou, Alex Andonian, Aude Oliva, and Antonio Torralba.
\newblock Temporal relational reasoning in videos.
\newblock In {\em Computer Vision - {ECCV} 2018 - 15th European Conference,
  Munich, Germany, September 8-14, 2018, Proceedings, Part {I}}, pages
  831--846, 2018.

\end{thebibliography}
}

\appendix
\section*{Appendix}
\section{Action classification networks}
In the main paper, we provide an overview of the gains in accuracy and speedup enabled by SCSampler for several video-classification models. In this section, we provide the details of the action classifier architectures used in our experiments and discuss the training procedure used to train these models. 

\subsection{Architecture details}
3D-ResNets (R3D) are residual networks where every convolution is 3D. Mixed-convolution models (MC$x$) are 3D CNNs leveraging residual blocks, where the first $x-1$ convolutional groups use 3D convolutions and the subsequent ones use 2d convolutions. In our experiments we use an MC3 model. R(2+1)D are models that decompose each 3D convolution in a 2D convolution (spatial), followed by 1D convolution (temporal). For further details, please refer to the paper that introduced and compared these models~\cite{tran2017closer} or the repository~\cite{VMZ} where pretrained models can be found.

\subsection{Training procedure}

\noindent\textbf{Sports-1M.}
For the Sports1M dataset, we use the training procedure described in~\cite{tran2017closer} for all models except ip-CSN-152. Frames are first re-scaled to have resolution $342 \times 256$, and then each clip is generated by randomly cropping a window of size $224 \times 224$ at the same location from 16 adjacent frames.  We use batch normalization after all convolutional layers, with a batch size of 8 clips per GPU. The models are trained for 100 epochs, with the first 15 epochs used for warm-up during distributed training. Learning rate is set to 0.005 and divided by 10 every 20 epochs. The ip-CSN-152 model is trained according to the training procedure described in \cite{Tran19}. \\

\noindent\textbf{Kinetics.}
On Kinetics, the clip classifiers are trained with mini-batches formed by sampling five 16-frame clips with temporal jittering. Frames are first resized to resolution $342 \times 256$, and then each clip is generated by randomly cropping a window of size $224 \times 224$ at the same location from 16 adjacent frames. The models are trained for 45 epochs, with 10 warm-up epochs. The learning rate is set to 0.01 and divided by 10 every 10 epochs as in~\cite{tran2017closer}. ip-CSN-152~\cite{Tran19} and R(2+1)D~\cite{tran2017closer} are finetuned from Sports1M for 14 epochs with the procedure described in~\cite{Tran19}.


\section{Implementation details for SCSampler}
In this section, we give the implementation details of the architectures and describe the training/finetuning procedures of our sampler networks.\\

\subsection{Visual-based sampler}
 Following Wu et al.~\cite{wu2018coviar}, all of our visual samplers are pre-trained on the ILSVRC dataset~\cite{ILSVRC15}. The learning rate is set to 0.001 for both Sports1M and Kinetics. As in~\cite{wu2018coviar}, the learning rate is reduced when accuracy plateaus and pre-trained layers use $100\times$ smaller learning rates. The ShuffleNet0.5~\cite{zhang2018shufflenet} (26 layers) model is pretrained on ImageNet. We use three groups of group convolutions as this choice is shown to give the best accuracy in~\cite{zhang2018shufflenet}. The initial learning rate and the learning rate schedule are the same as those used for ResNet-18.\\

\subsection{Audio-based sampler}
We use a VGG model~\cite{vgg2014} pretrained on AudioSet~\cite{audioset} as our backbone network, with MEL spectrograms of size $40\times200$ as input. When fine-tuning the network with \textit{SAL-RANK}, we use an initial learning rate of 0.01 for Sports1M and 0.03 for Kinetics for the first 5 epochs and then divide the learning rate by $10$ every 5 epochs. The learning rate of the pretrained layers is multiplied by a factor of $5*10^{-2}$. When finetuning with the \textit{SAL-CL} loss, we set the learning rate to 0.001 for 10 epochs, and divide it by 10 for  6 additional epochs. When finetuning with \textit{AC} loss, we start with learning rate 0.001, and divide it by 10 every 5 epochs.

\section{Additional evaluations of design choices for SCSampler}

Here we present additional analyses of the design choices and hyperparameter values of SCSampler.

\subsection{Varying the audio sampler architecture.}

Table~\ref{tab:audio_features} shows video classification accuracy using different variants of our audio sampler. Given our VGG audio network pretrained for classification on AudioSet, we train it on miniSport using the following two options: finetuning the entire VGG model  vs training a single FC layer on VGG activations from one layer (conv4\_2, pool4, or fc1). 
All audio samplers are trained with the SAL-RANK loss. We can see that finetuning the audio sampler gives the best classification accuracy. 

\begin{table}[]
\captionsetup{font=small}
\footnotesize
\centering
\begin{tabular}{@{}ccc@{}}
\toprule
Audio SCSampler & accuracy (\%) & runtime (min) \\ \midrule
\multicolumn{1}{c|}{finetuned VGG} & 67.82 & 22.0 \\
\multicolumn{1}{c|}{FC trained on VGG-conv4\_2} & 67.03 & 21.6 \\
\multicolumn{1}{c|}{FC trained on VGG-pool4} & 67.01 & 21.4 \\
\multicolumn{1}{c|}{FC trained on VGG-fc1} & 59.84 & 21.4 \\
\bottomrule
\end{tabular}
\vspace{-6pt}
\caption{\small Varying the audio sampler architecture. 
Performance is measured as MC3-18 video accuracy (\%) on the test set of miniSports with $K=10$ sampled clips.}
\label{tab:audio_features}
\end{table}

\subsection{Varying the number of sampled clips ($K$)}

\begin{figure}[t]
\captionsetup{font=small}
\begin{center}
   \includegraphics[width=0.9\linewidth]{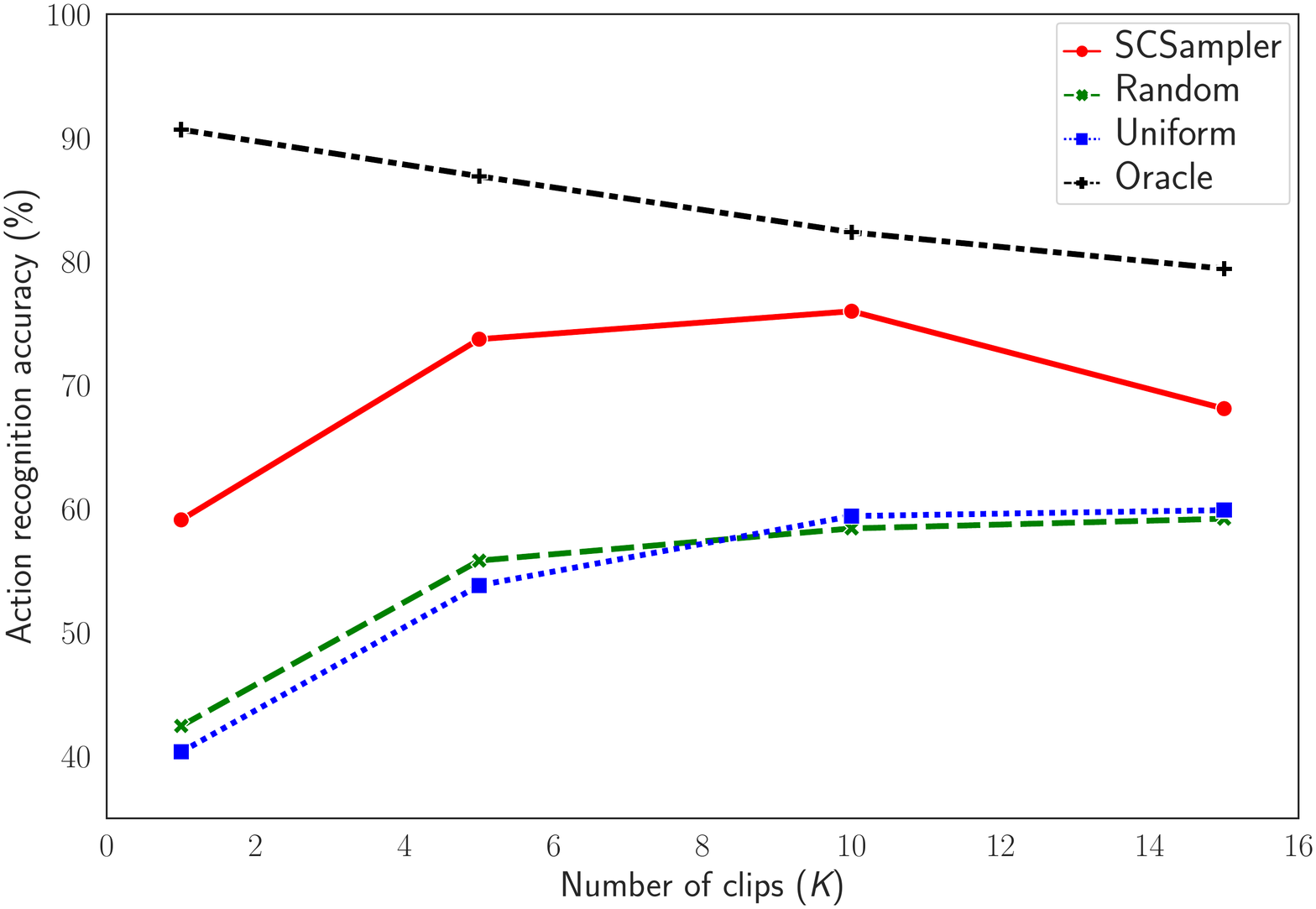}
\end{center}
\vspace{-15pt}
   \caption{\small Video classification accuracy (\%) of MC3-18 on the miniSports test set vs the number of sampled clips ($K$).}
\label{fig:abl_k}
\end{figure}

Figure~\ref{fig:abl_k} shows how video-level classification accuracy changes as we vary the number of sampled clips ($K$).
The sampler here is {\em AV-union-list}. 
$K=10$ provides the best accuracy for our sampler. For the Oracle, $K=1$ gives the top result as this method can conveniently select the clip that elicits the highest score for the correct label on each test video.  

\subsection{Selecting hyperparameter $K'$ for \textit{AV-union-list}}
\begin{figure}[t]
\begin{center}
\includegraphics[width=\linewidth]{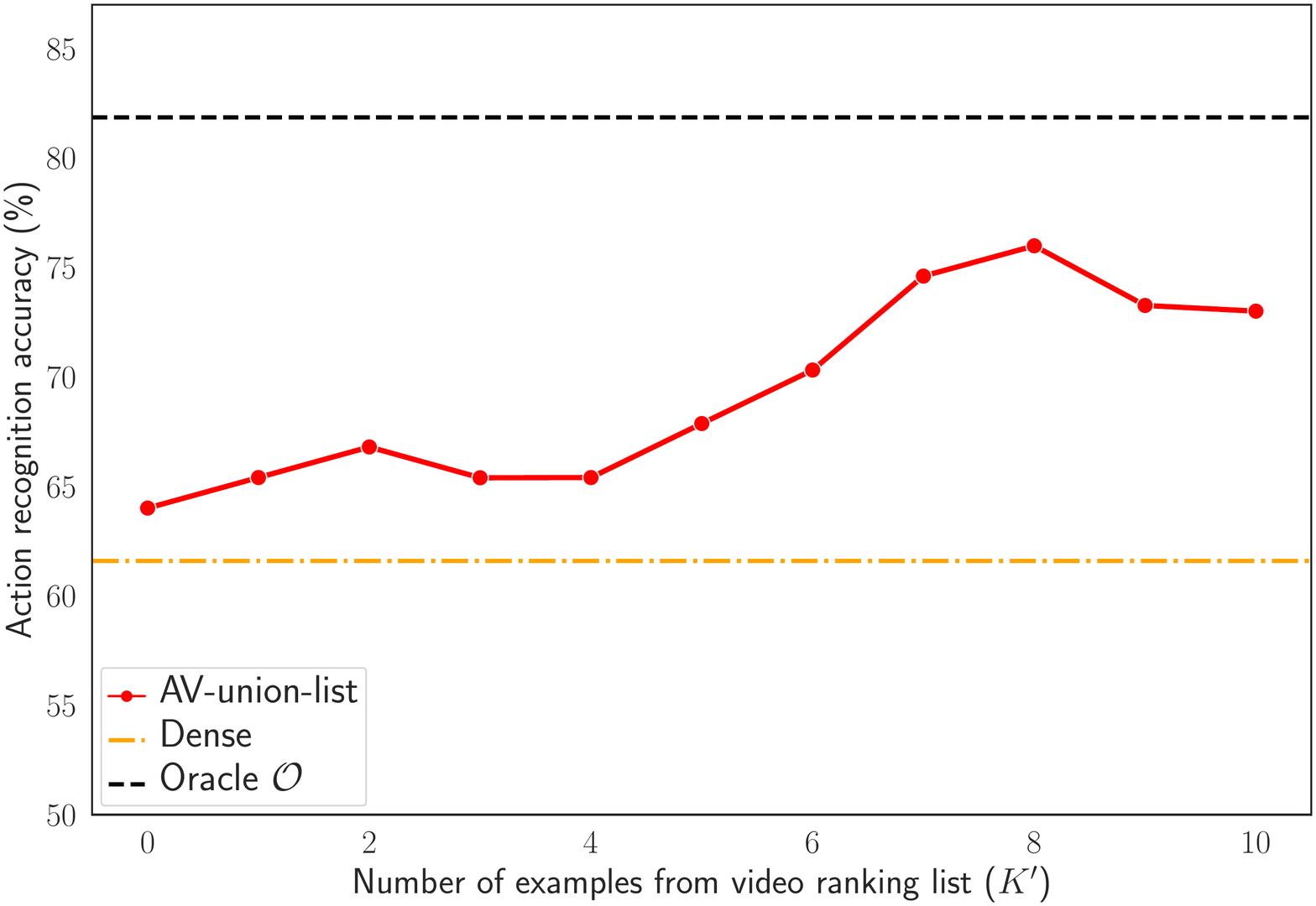}
\end{center}
   \caption{Varying the number of clips $K'$ sampled by the visual sampler, when combining video-based and and audio-based sampler according to the \textit{AV-union-list} strategy. The best action recognition accuracy is achieved when sampling $K'=8$ clips with the video-based sampled and $K-K'=2$ clips with the audio-based sampler. Evaluation is done on the miniSports dataset, with the MC3-18 clip classifier.}
\label{fig:Kprime}
\end{figure}

The \textit{AV-union-list} method (described in section 3.3.3 of our paper) combines the audio-based and the video-based samplers, by selecting $K'$ top-clips according to the visual sampler (with hyper-parameter $K'$ s.t. $K'<K$) and adds a set of $K-K'$ different clips from the ranked list of the audio sampler to form a sample set of size $K$ ($K=10$ is used in this experiment). In Figure \ref{fig:Kprime} we analyze the impact of $K'$ on action classification. The fact that the best value is achieved at $K'=8$ suggests that the signals from the two samplers are somewhat complementary, but the visual sampler provides a more accurate measure of clip saliency.

\section{Comparison to Random/Uniform under the same runtime.} 
\begin{figure}[t]
\begin{center}
\includegraphics[width=\linewidth]{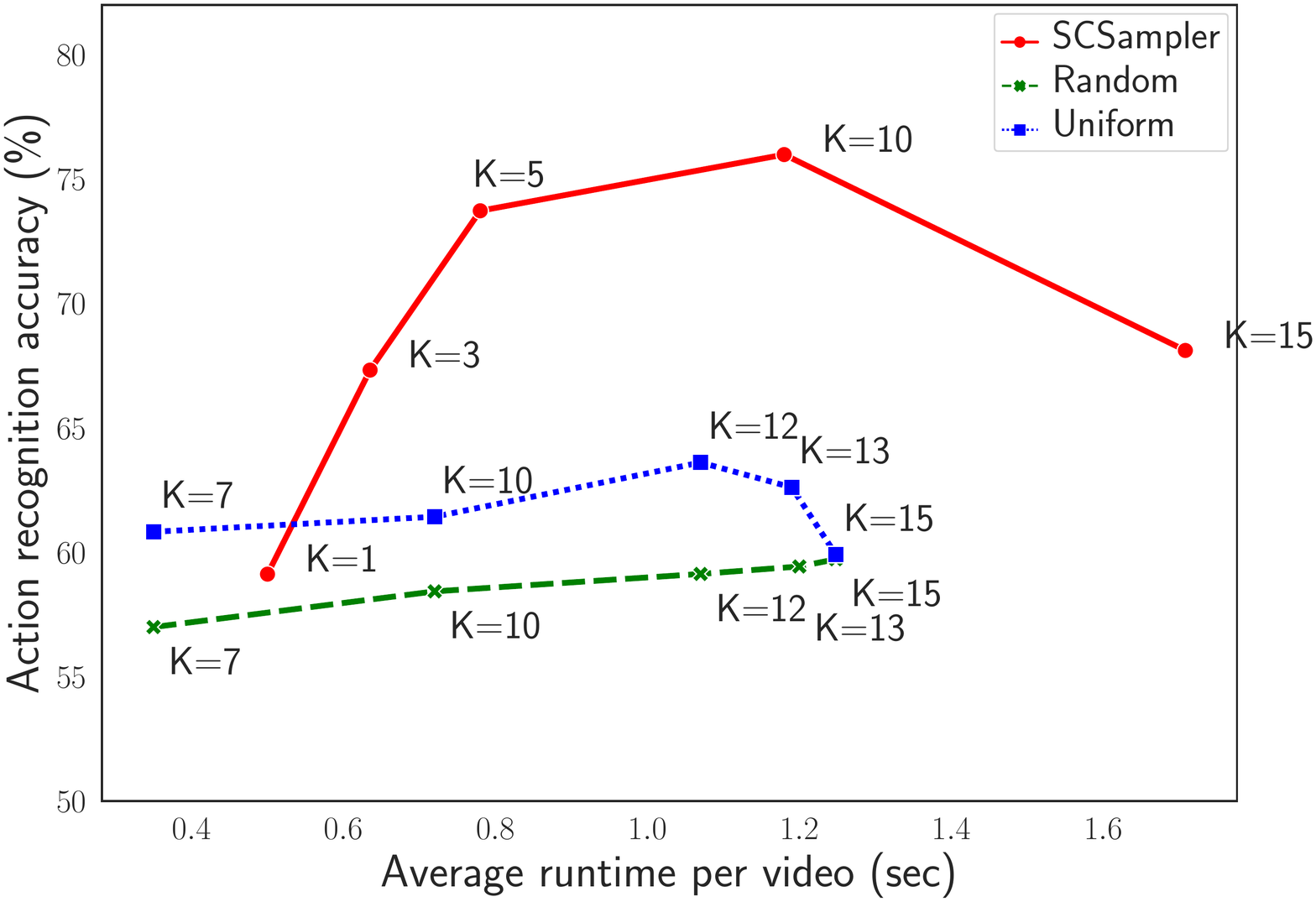}
\end{center}
\caption{\small Video-level classification accuracy on the test of miniSports vs runtime per video using different numbers of sampled clips ($K$). The clip classifier is MC3-18.}
\label{fig:SameRuntime}
\end{figure}

Fig.~\ref{fig:SameRuntime} shows runtime (per video) vs video-level classification accuracy on miniSports, obtained by varying the number of sampled clips per video ($K$). For this test we use MC3-18, which is the fastest clip-classifier in our comparison. The overhead of running SCSampler on each video is roughly equivalent to 3 clip-evaluations of MC3-18. Even after adding clip evaluations to Random/Uniform to obtain a comparison under the same runtime, SCSampler significantly outperforms these baselines. Note that for costlier clip-classifiers the SCSampler overhead would amount to less than one clip evaluation (e.g., 0.972 for R(2+1)D-50), making the option of Random/Uniform even less appealing for the same runtime. 
 
\section{Applying SCSampler every $N$ clips}
\begin{figure}[h]
\includegraphics[width=\linewidth]{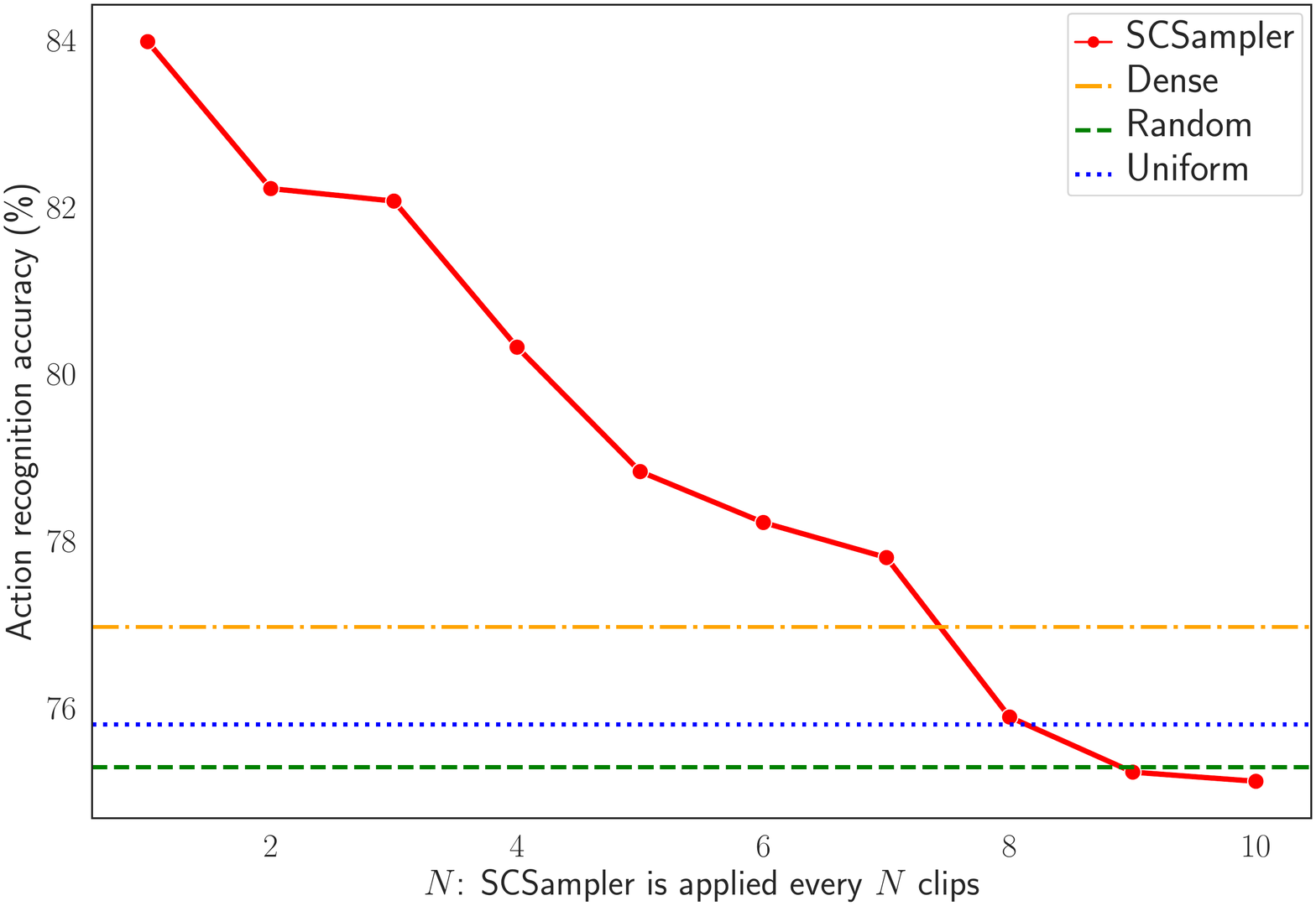}

   \caption{Applying SCSampler every $N$ clips reduces the computational cost. Here we study how applying SCSampler with a clip-stride of $N$ affects the action classification accuracy on Sports1M using ip-CSN-152 as clip classifier.}
\label{fig:n}
\end{figure}

While our sampler is quite efficient, further reductions in computational cost can be obtained by running SCSampler every $N$ clips in the video. This implies that the final top-$K$ clips used by the action classifier will be selected from a subset of clips obtained by applying SCSampler with a stride of $N$ clips. As usual, we fix the value of $K$ to 10 for SCSampler. Figure~\ref{fig:n} shows the results obtained with the best configuration of our SCSampler (see details in 4.1.1) and the ip-CSN-152~\cite{Tran19} action classifier on the full Sports1M dataset. We see that we can apply SCSampler with clip-strides of up to $N=7$ before the action recognition accuracy degrades to the level of costly dense predictions. This results in further reduction of computational complexity and runtime, as we only need to apply the sampler to $\lceil L/N \rceil$ clips. 

\end{document}